\documentclass{article}
\usepackage{ijcai21}

\usepackage{times}
\usepackage{soul}
\usepackage{url}
\usepackage[utf8]{inputenc}
\usepackage[small]{caption}
\usepackage{graphicx}
\usepackage{amsmath}
\usepackage{amsthm}
\usepackage{booktabs}
\usepackage{algorithm}
\usepackage{algorithmic}
\usepackage{subfigure}
\usepackage{siunitx}
\usepackage{stfloats}
\usepackage{multicol}  
\usepackage{multirow} 
\usepackage{makecell}
\usepackage{booktabs}
\usepackage{lipsum}
\newcommand{\RNum}[1]{\lowercase\expandafter{\romannumeral #1\relax}}

\title{Multi-view Feature Augmentation with Adaptive Class Activation Mapping}

\iftrue
\author{
Xiang Gao$^{1,2,3}$\
Yingjie Tian$^{2,3}$\and
Zhiquan Qi$^{2,3}$\\
\affiliations
$^1$School of Computer Science and Technology, University of Chinese Academy of Sciences\\
$^2$Research Center on Fictitious Economy and Data Science, Chinese Academy of Sciences\\
$^3$Key Laboratory of Big Data Mining and Knowledge Management, Chinese Academy of Sciences\\
\emails
gaoxiang181@mails.ucas.ac.cn,
tyj@ucas.ac.cn,
qizhiquan@foxmail.com
}
\fi

\begin{document}

\maketitle

\begin{abstract}
  We propose an end-to-end-trainable feature augmentation module built for image classification that extracts and exploits multi-view local features to boost model performance. Different from using global average pooling (GAP) to extract vectorized features from only the global view, we propose to sample and ensemble diverse multi-view local features to improve model robustness. To sample class-representative local features, we incorporate a simple auxiliary classifier head (comprising only one 1$\times$1 convolutional layer) which efficiently and adaptively attends to class-discriminative local regions of feature maps via our proposed AdaCAM (Adaptive Class Activation Mapping). Extensive experiments demonstrate consistent and noticeable performance gains achieved by our multi-view feature augmentation module. 
\end{abstract}

\newcommand\blfootnote[1]{
    \begingroup
    \renewcommand\thefootnote{}\footnote{#1}
    \addtocounter{footnote}{-1}
    \endgroup
}
\blfootnote{\textbf{Accepted by IJCAI 2021}}

\section{Introduction}
Progresses of image classification were dramatically promoted by deep convolutional neural networks. NIN \cite{NIN} for the first time replaced traditional flattening-based fully-connected (FC) layers with global average pooling (GAP), which reduces model size and enables input images of arbitrary size. Since then, GAP is widely used in combination with more complex convolutional backbones. GoogLeNet \cite{googlenet} extracts and fuses multi-scale convolutional features to enrich feature representation. ResNet \cite{resnet} introduces residual learning to facilitate optimization of deep networks, improving model performance noticeably. Later on, more variants and derivatives of ResNet are proposed to make further improvements by increasing network width \cite{wideresnet}, adopting multi-path group convolution \cite{ResNeXt}, extending skip connection to dense connections \cite{huang2017densely}, and introducing attention mechanism \cite{wang2017residual,zhang2020resnest}. Besides, networks were also specially designed for lightweight deployment by using more efficient convolution or convolutional blocks \cite{mobilenets,squeezenet}. 
\begin{figure}[t]
\includegraphics[width=3.25in,height=1.5in]{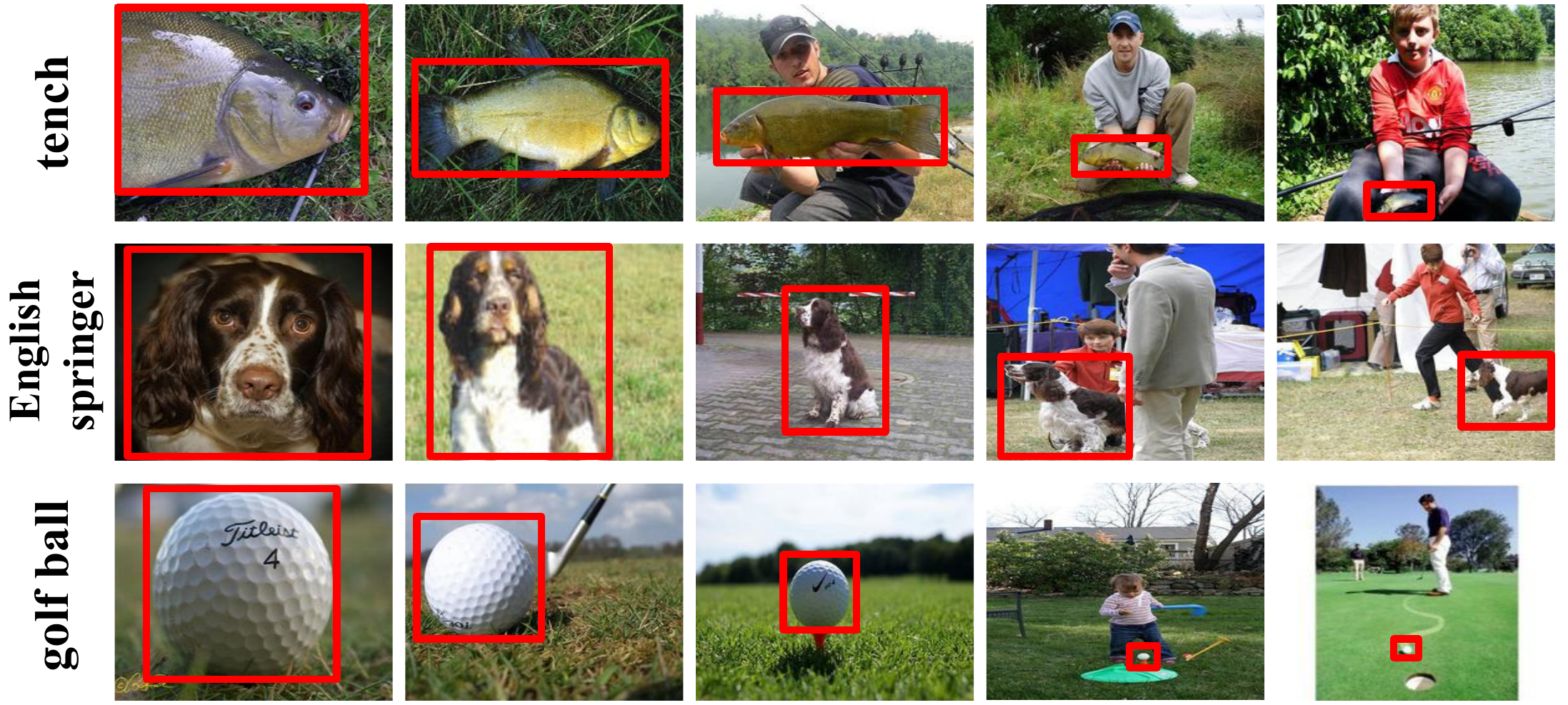}
\caption{Example images of ``tench'', ``English springer'', ``golf ball'' categories in ImageNet dataset. Even for images of the same class, the class-related objects, which we annotate with red boxes, vary a lot in scale. We consider that for an image with a small scale of class-related object, the final image representation extracted by global average pooling (GAP) could be corrupted by class-irrelevant background features, and thus is less representative of the corresponding class. This motivates us to attend to local region of class-related object and extract more class-representative image representations by sampling local features around the attended region.}
\label{motivation}
\end{figure}
However, all these advanced deep models only focus on delicate design of convolutional backbone networks, the common ground is applying GAP to obtain vectorized features which are then fed to a linear classifier head for final classification. Though effective and efficient, GAP extracts vectorized image representations simply by averaging cross the entire feature maps. In this way, the extracted image representations could be corrupted by background elements for images where the class-related object takes up only a fraction of the image. As Figure \ref{motivation} displays, even for images of the same class, the scale of class-related object varies a lot in ImageNet dataset. For images with a small scale of class-related object, GAP easily extracts image representations that bias towards features of the background or class-irrelevant objects, making the input features to the subsequent linear classifier head less class-representative. 
% Intuitively, it could be less convincing to categorize an image with an English springer appearing only at a small region into the ``English springer'' class from the global view of the image.

\begin{figure}[t]
\includegraphics[width=3.3in]{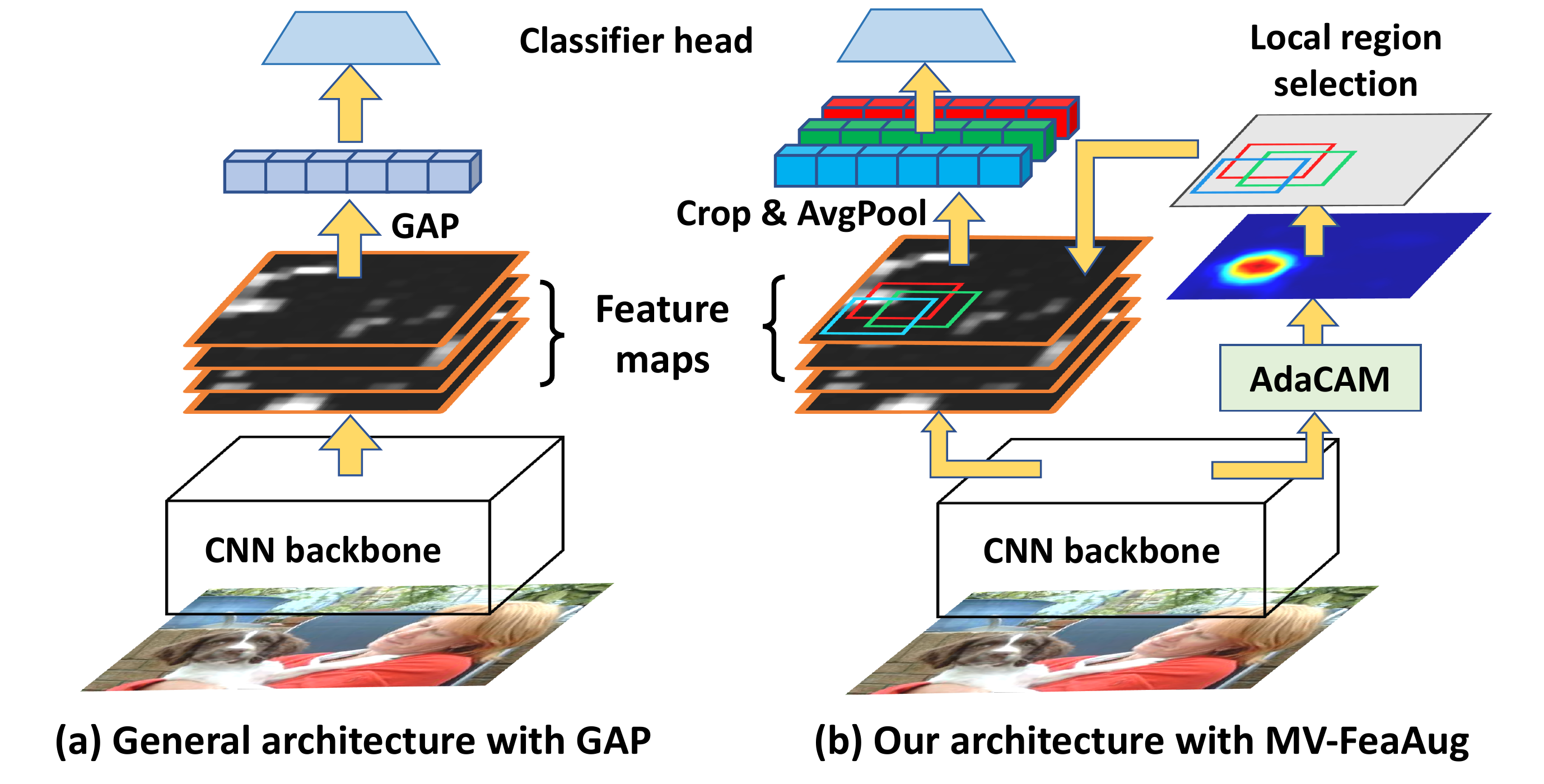}
\caption{Comparison between the image classification architecture with the general GAP (left) and our MV-FeaAug module (right). We sample diverse local features around the class-discriminative region of the final convolutional feature maps as multi-view local image representations for ensembled classification, as compared with GAP that extracts only global-view image representation.}
\label{comparison_to_GAP}
\end{figure}

Based on this motivation, we replace GAP with a multi-view feature augmentation module (MV-FeaAug), which learns to locate region of class-related object on the final convolutional feature maps, and extracts multi-view image representations by sampling diverse local features around the attended class-discriminative region. The extracted multi-view local representations more precisely represent the class-related object of the image from diverse views, the ensemble of the predictions to all these representations are used as the final prediction of the entire image. Comparison between the image classification architecture with the general GAP and our MV-FeaAug counterpart is illustrated in Figure \ref{comparison_to_GAP}. 

Apart from network architecture design, data augmentation, which acts as a regularizer to reduce overfitting, is also important for image classification. In addition to the basic position and color transformations, some advanced methods were proposed in recent years. Mixup \cite{mixup} blends two images and their corresponding labels with a random proportion as training samples. Cutout \cite{cutout} randomly removes a rectangular patch from images to guide the network to learn from local views. CutMix \cite{cutmix} randomly cuts and pastes rectangular patches among training images with their corresponding labels mixed up in proportion to patch area. Our method shares similar spirit in expanding training samples with these methods. But instead of image-level augmentation, our method extracts diverse multi-view local representations at feature space as augmented samples for subsequent classifier head, for which we call feature augmentation. Besides, our method takes full advantage of the augmented features at inference time by ensembling predictions to all the multi-view features, while such ensemble prediction property is not available for the above-mentioned data augmentation methods.

Our method also relates to ``visual explanation'' of CNN models. Efforts have been made to visualize the evidence of class predictions made by CNN. CAM \cite{CAM} produces class activation maps that highlight class-discriminative regions of images with GAP and a single-layer classifier head. Grad-CAM \cite{Grad-CAM} extends such object localization ability to arbitrary CNN architectures by computing gradients of the logits with respect to intermediate feature maps. However, both CAM and Grad-CAM operate in a way of postprocessing since they require extraction (or computation) of weights (or gradients) associated to the target-class logit. Moreover, the attention maps produced by these two methods are conditioned on the ground truth label, which is not flexible in cases when the target image label is unknown. In this work, we propose AdaCAM that adaptively generates attention maps with simply a forward pass of the network without conditioning on the class label. The produced attention maps provide cues to sample diverse class-representative local features for feature augmentation.

\begin{figure}[t]
\includegraphics[width=3.35in]{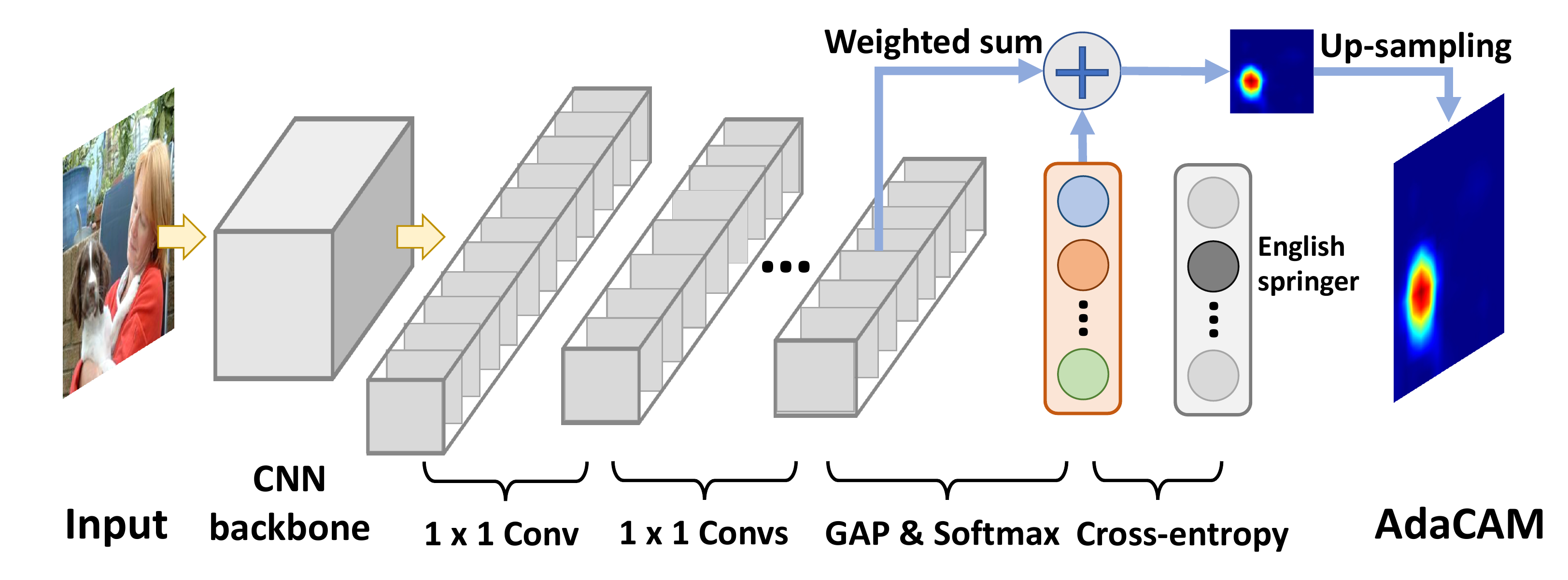}
\caption{Adaptive class activation mapping (AdaCAM). We replace the traditional classifier head made up of [GAP$\rightarrow$MLP$\rightarrow$Softmax] with [MLPConv$\rightarrow$GAP$\rightarrow$Softmax] (MLPConv comprises consecutive Conv$_{1\times1}$ layers joined by non-linear activations) to maintain spatial resolution of feature maps. The AdaCAM is obtained by performing channel-wise weighted sum of the last convolutional feature maps with respect to the softmax logit vector.}
\label{AdaCAM}
\end{figure}

\begin{figure*}[t]
\centering
\includegraphics[width=0.93\textwidth]{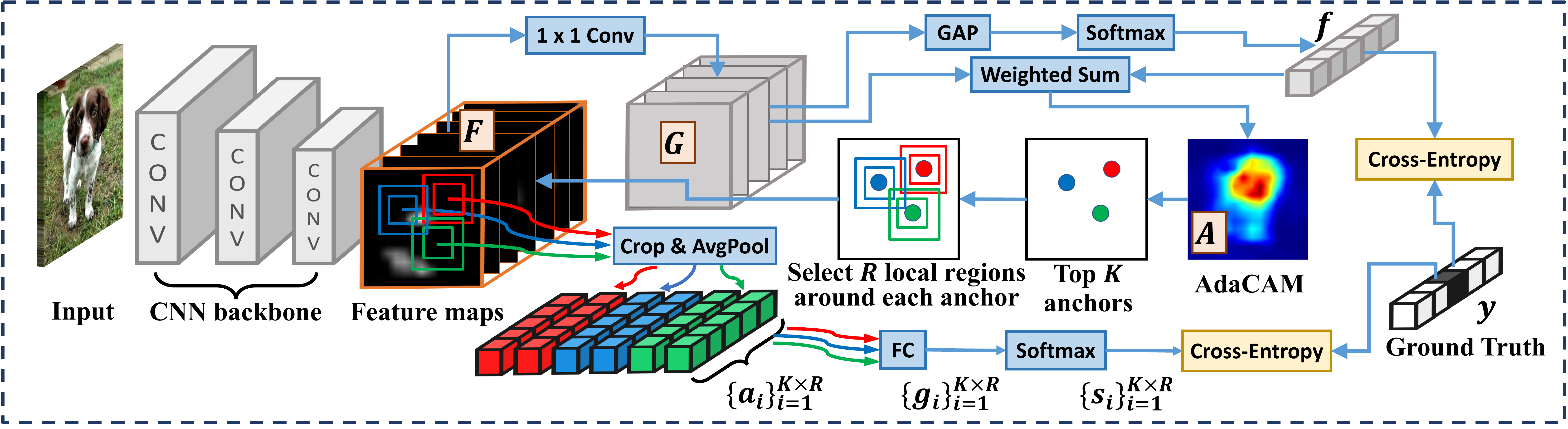}
\caption{Overview of MV-FeaAug. We concurrently train an auxiliary classifier head comprised of only one 1$\times$1 convolutional layer for dynamic generation of AdaCAM, based on which we sample multiple local representations on the final convolutional feature maps as multi-view inputs to the main classifier head. The main classifier head comprises (but not restricted to) a single fully-connected layer.}
\label{overview}
\end{figure*}

Another topic related to our method is attention mechanism, which has been widely applied to image classification recently. SENet \cite{SENet} built a Squeeze-and-Excitation module to selectively enhance or depress features in channel dimension. CBAM \cite{cbam} sequentially stacked two modules for channel-wise and spatial attentions respectively. SKNet \cite{SKNet} proposed a multi-branch attention module for adaptive convolution kernel size selection. GCNet \cite{gcnet} improved model performance by designing an efficient self-attention block that captures global context information. All these methods increase model size for incorporating additional attention modules. By contrast, our method also introduces attention mechanism by locating class-discriminative regions via AdaCAM, however, all the extra cost is an auxiliary classifier head comprised of only one 1$\times$1 convolutional (Conv$_{1\times1}$) layer. Furthermore, we demonstrate that a classification network with a deeper CNN backbone could be surpassed by a network with a shallower backbone combined with our MV-FeaAug module.

\section{Method}
In this section, we firstly describe our method for efficiently producing label-independent attention maps, i.e., \textit{adaptive class activation mapping} (AdaCAM), and then elaborate on the complete architecture of our MV-FeaAug module. 

\subsection{Adaptive Class Activation Mapping}
We start with a brief introduction of CAM and Grad-CAM for completeness and ease of comparison.

\textbf{CAM} uses GAP and a single-FC-layer classifier head to produce class activation maps. Supposing the output feature maps of the CNN backbone are $\left\{F_k\right\}_{k=1}^{K}$, where $F_k$ is the $k$-th channel, and $\left\{u_k\right\}_{k=1}^{K}$ are the units after GAP, i.e., $u_k=\frac{1}{HW}\sum_{x,y}{F_{k}(x,y)}$, where $H$ and $W$ denote the height and width of feature maps. The class activation map for class $c$ is $A_{c} = \sum_{k}{w_{k}^{c}F_{k}}$, where $w_{k}^{c}$ is the weight of the subsequent FC layer that corresponds to class $c$ for unit $u_{k}$. The limitation of CAM is threefold: (\RNum{1}) it is less efficient for the need of extracting intermediate weights of the model; (\RNum{2}) it is limited to single-layer classifier head, and thus may compromise classification performance as compared with MLP classifier head; (\RNum{3}) it is conditioned on the ground truth label, which may not be known in practical applications.

\textbf{Grad-CAM} addresses the second limitation of CAM by backpropagating gradients with respect to convolutional feature maps. Specifically, the class activation maps for class $c$ is $A_{c}=\sum_{k}{\alpha_{k}^{c}F_{k}}$, in which $\alpha_{k}^{c}$ is the gradient of $f^{c}$, the logit (before softmax) of class $c$, with respect to $F_{k}$, i.e., $\alpha_{k}^{c}=\frac{1}{HW}\sum_{x,y}{\partial{f^{c}}\big/\partial{F_{k}(x,y)}}$. However, Grad-CAM is still not efficient for the need of gradient computation, and also relies on the specific class label.

A key ingredient of our MV-FeaAug module is to employ class activation maps to highlight class-related regions for local features sampling, for which a good method to produce such attention maps should follow two criteria: (\RNum{1}) the generation process is simple and efficient; (\RNum{2}) the generation process is independent of class label, since the ground truth labels are not given at inference time. Both CAM and Grad-CAM fail to satisfy these two conditions, for which we propose AdaCAM to meet the requirements.

\textbf{AdaCAM} aims to directly synthesize class activation maps in a feed-forward pass without conditioning on target labels. This functionality is achieved with only a slight modification of the classifier head. As illustrated in Figure \ref{AdaCAM}, after feature extraction of CNN backbone, we alter the traditional classifier head structure of [GAP$\rightarrow$MLP$\rightarrow$ Softmax] into [MLPConv$\rightarrow$GAP$\rightarrow$Softmax], where MLPConv refers to consecutive Conv$_{1\times1}$ layers joined by non-linear activations. The Conv$_{1\times1}$ layers are essentially equivalent to FC layers, but have the advantage of maintaining spatial resolution of feature maps. The output of MLPConv is feature maps $\{G_{c}\}_{c=1}^{C}$ with its channel number equalling the number of classes $C$, i.e., $\{G_c\}_{c=1}^{C}$$=$MLPConv($\{F_k\}_{k=1}^{K}$), where $G_{c}$ represents the $c$-th channel. Then GAP and softmax operations are employed to obtain the final softmax logit vector $\textbf{\textit{f}}$, i.e., $\textbf{\textit{f}}$$=$Softmax(GAP($\{G_c\}_{c=1}^{C}$)), the objective is to minimize the cross-entropy between $\textbf{\textit{f}}$ and the one-hot label vector $\textbf{\textit{y}}$. This architecture is equivalent to traditional MLP classifier head since MLPConv shares the same parameter space with MLP. The advantage is the efficiency in obtaining class activation maps: for an image of class $c^\prime$, the corresponding class activation map $A_{c^\prime}$ is exactly $G_{c^\prime}$, the $c^\prime$-th channel of the MLPConv output (we prove this in supplementary materials). To further make the attention map independent of class label, we perform weighted sum on $\left\{G_c\right\}_{c=1}^{C}$ with respect to the softmax logit vector $\textbf{\textit{f}}$ as an approximation, i.e., $A=\sum_{i=1}^{C}{G_{i}f_{i}}$, where $f_{i}$ is the $i$-th element of $\textbf{\textit{f}}$. Actually, $A$ is very close to $A_{c^\prime}$ since softmax operation exponentially widens the gap between elements of the logit vector.

In this work, we build a separate auxiliary classifier head tailored for AdaCAM to provide attention maps for the main classifier head. Since the auxiliary classifier is not necessarily to be sufficiently accurate, we use only one Conv$_{1\times1}$ layer in the MLPConv part to reduce additional overhead.

\subsection{Multi-view Feature Augmentation}
The overall architecture of MV-FeaAug is illustrated in Figure \ref{overview}. An auxiliary classifier head is built on top of $\textbf{\textit{F}}=\left\{F_{k}\right\}_{k=1}^{K}$, the final feature maps of CNN backbone, to dynamically generate AdaCAM attention maps, based on which we sample multi-view local representations from $\textbf{\textit{F}}$ as input to the main classifier head. The final prediction at inference time is the ensemble of the predictions made by the main classifier head to all the sampled multi-view representations.

In the auxiliary classifier part, we first use a MLPConv network comprised of a single Conv$_{1\times1}$ layer to reduce the depth of feature maps to the class number $C$, i.e., $\textbf{\textit{G}}=\{G_{c}\}_{c=1}^{C}$$=$Conv$_{1\times1}$($\textbf{\textit{F}}$), then the softmax logit vector $\textit{\textbf{f}}$ is obtained after GAP and softmax operations, i.e., $\textbf{\textit{f}}$$=$Softmax(GAP($\textbf{\textit{G}}$)), the objective function is the cross-entropy between $\textit{\textbf{f}}$ and the one-hot label vector $\textbf{\textit{y}}$:
\begin{equation}
L_{global}=-\sum\nolimits_{i=1}^{C}y_{i}log(f_{i}),
\label{global_loss}
\end{equation}
where $y_{i}$ and $f_{i}$ are the $i$-th element of $\textbf{\textit{y}}$ and $\textbf{\textit{f}}$ respectively. We call this global loss as our auxiliary classifier considers only the global view of image features by using GAP. The minimization of the global loss endows the auxiliary classifier head object localization ability, which allows for dynamic generation of AdaCAM attention map $A$:
\begin{equation}
A=\sum\nolimits_{i=1}^{C}G_{i}f_{i}.
\label{AdaCAM}
\end{equation}
The attention map $A$ shares the same spatial size as feature maps $\textbf{\textit{F}}$, and highlights class-discriminative local regions of the image. The pixel intensity of $A$ reflects the correlation of each location with the corresponding image class. Therefore, we select top $K$ locations with the largest pixel values from $A$ as anchor points, around which we sample multiple local regions from $\textbf{\textit{F}}$ for feature augmentation. Let $V$ be the set of anchor points for an input image, $S$ be an ordered list of the pixel coordinates of $A$: 
\begin{equation}
\begin{aligned}
&S=\{(x_{i},y_{i}) | i=1,2,...,HW, \\
&A(x_{1},y_{1})\ge A(x_{2},y_{2})\ge...\ge A(x_{HW},y_{HW})\},
\end{aligned}
\end{equation}
where $H$ and $W$ denote the height and width of $A$, then the anchor set $V$ is the top $K$ subset of S:
\begin{equation}
V=\{(x_{i},y_{i}) | (x_{i},y_{i})\in S,i=1,2,...,K\}.
\end{equation}
The number of anchor points $K$ for each input image is a hyper-parameter of the model. Each anchor point in $V$ serves as a centroid around which we crop $R$ multi-scale square regions on $\textbf{\textit{F}}$ for feature augmentation. For feature maps $\textbf{\textit{F}}$ with spatial size $H\times W$, we pre-define a region size list $L_{R}=[r_{1}, r_{2}, ..., r_{R}]$, in which $r_{i}$ (odd number) denotes the length of the side of the $i$-th square region to be cropped around each anchor point in $V$. Specifically, given a centroid $(x_{c},y_{c})\in V$ and a region length $r \in L_{R}$, the square region to be cropped on $\textbf{\textit{F}}$ is determined by the coordinates of the top-left corner $(x_{tl}, y_{tl})$ and the bottom-right corner $(x_{br}, y_{br})$:
\begin{equation}
\left\{
      \begin{array}{lr}
        x_{tl}=max(x_{c}-(r-1)/2, 0)   \\
        y_{tl}=max(y_{c}-(r-1)/2, 0)   \\
        x_{br}=min(x_{c}+(r-1)/2, W-1) \\
        y_{br}=min(y_{c}+(r-1)/2, H-1). \\
      \end{array}
\right.
\end{equation}
The length and the elements of $L_{R}$ are also hyper-parameters of our method. The purpose of sampling multi-scale local regions is to (\RNum{1}) capture class-related objects with different and varying spatial sizes, (\RNum{2}) further increase the number of the augmented features to reduce overfitting, (\RNum{3}) extract features of the class-related objects from more views to improve classification robustness. In this way, there are $K\times R$ local feature maps cropped out from $\textbf{\textit{F}}$ for each input image, to which average pooling is applied separately, yielding the augmented vectorized features $\{\textbf{\textit{a}}_{i}\}_{i=1}^{K\times R}$ as multi-view inputs to the main classifier head, as compared with the traditional single-view input GAP(\textbf{\textit{F}}). Like most classic networks \cite{resnet,googlenet,ResNeXt,mobilenets}, the main classifier head comprises a single FC layer that converts the vectorized features into logit vectors $\{\textbf{\textit{g}}_{i}\}_{i=1}^{K\times R}$, i.e., $\{\textbf{\textit{g}}_{i}\}_{i=1}^{K\times R}$$=$FC($\{\textbf{\textit{a}}_{i}\}_{i=1}^{K\times R}$), after which softmax is applied to obtain probability distribution vectors $\{\textbf{\textit{s}}_{i}\}_{i=1}^{K\times R}$, i.e., $\{\textbf{\textit{s}}_{i}\}_{i=1}^{K\times R}$$=$Softmax($\{\textbf{\textit{g}}_{i}\}_{i=1}^{K\times R}$). All the augmented multi-view features are classified to the corresponding image-level label by minimizing the cross-entropy between $\{\textbf{\textit{s}}_{i}\}_{i=1}^{K\times R}$ and the one-hot label vector $\textbf{\textit{y}}$:
\begin{equation}
L_{local}=-\sum\nolimits_{i=1}^{K\times R}\sum\nolimits_{c=1}^{C}y_{c}log(s_{i,c}),
\end{equation}
where $y_{c}$ and $s_{i,c}$ are the $c$-th element of $\textbf{\textit{y}}$ and $\textbf{\textit{s}}_{i}$ respectively. We call this local loss as the inputs to the main classifier head are multi-view local features. The total loss function is the combination of the global loss and the local loss:
\begin{equation}
L_{total}=L_{global}+L_{local}.
\label{total_loss}
\end{equation}

We find that the final classification performance is not sensitive to tuning the weights of $L_{global}$ and $L_{local}$ in Eq. \ref{total_loss}, and thus assign equal weights for the two loss functions.

At inference time, the final prediction of the image is the ensembled result of the predictions to all the augmented features, i.e., the predicted class label $p$ is:
\begin{equation}
p=\mathop{\arg\max}_{c}(\sum\nolimits_{i=1}^{K\times R}{g_{i,c}}).
\end{equation}

For computational efficiency, the inputs of the entire model is actually a mini-batch of $N$ images, the complete loss function including the batch dimension is as below:
\begin{equation}
\begin{aligned}
L_{total}=&-\frac{1}{N}\sum\nolimits_{n=1}^{N}\sum\nolimits_{c=1}^{C}{y_{c}^{(n)}log(f_{c}^{(n)})} \\
          &-\frac{1}{N}\sum\nolimits_{n=1}^{N}\sum\nolimits_{i=1}^{K\times R}\sum\nolimits_{c=1}^{C}y_{c}^{(n)}log(s_{i,c}^{(n)}),
\end{aligned}
\end{equation}
in which the superscript $(n)$ indicates the $n$-th image sample in a mini-batch. Last but not least, our MV-FeaAug module has no restriction on the main classifier head. Besides single-layer softmax classifier, the main classifier head could be more complex one with multiple hidden layers, or more robust one like prototype classifier \cite{yang2018robust}, we demonstrate this in supplementary materials. 
\section{Experiments}
\subsection{Experiment Setup}
We evaluate our method on the following object, scene, and action classification datasets: Caltech101, Imagenette, Corel5k, Scene15, MIT Indoor67, Stanford Action40, UIUC Event8, UCMLU, RSSCN7, and AID. The details of these datasets are described in supplementary materials. All the images of these datasets are resized to 224$\times$224. We verify the effectiveness of our MV-FeaAug module by applying it to networks with different CNN backbones. For all CNN backbones, the resolution of the output feature maps is kept to be 14$\times$14, upon which we sample 3$\times$3, 5$\times$5, 7$\times$7, and 9$\times$9 multi-scale local features around each anchor point, i.e., we set $L_{R}=[3,5,7,9]$. In all experiments, we use the Adam optimizer with $\beta_{1}=0.9$ and $\beta_{2}=0.999$. The initial learning rate is $lr=0.0001$, the batch size is $N=64$. We only use very simple data augmentation techniques by firstly flipping images horizontally with 0.5 probability, and then enlarging images to 256$\times$256 followed by randomly cropping back to 224$\times$224. Our model is implemented with TensorFlow and run on a single GeForce GTX 1080 Ti GPU.

\subsection{Experiment Results}

\begin{figure}[t]
\includegraphics[width=3.25in]{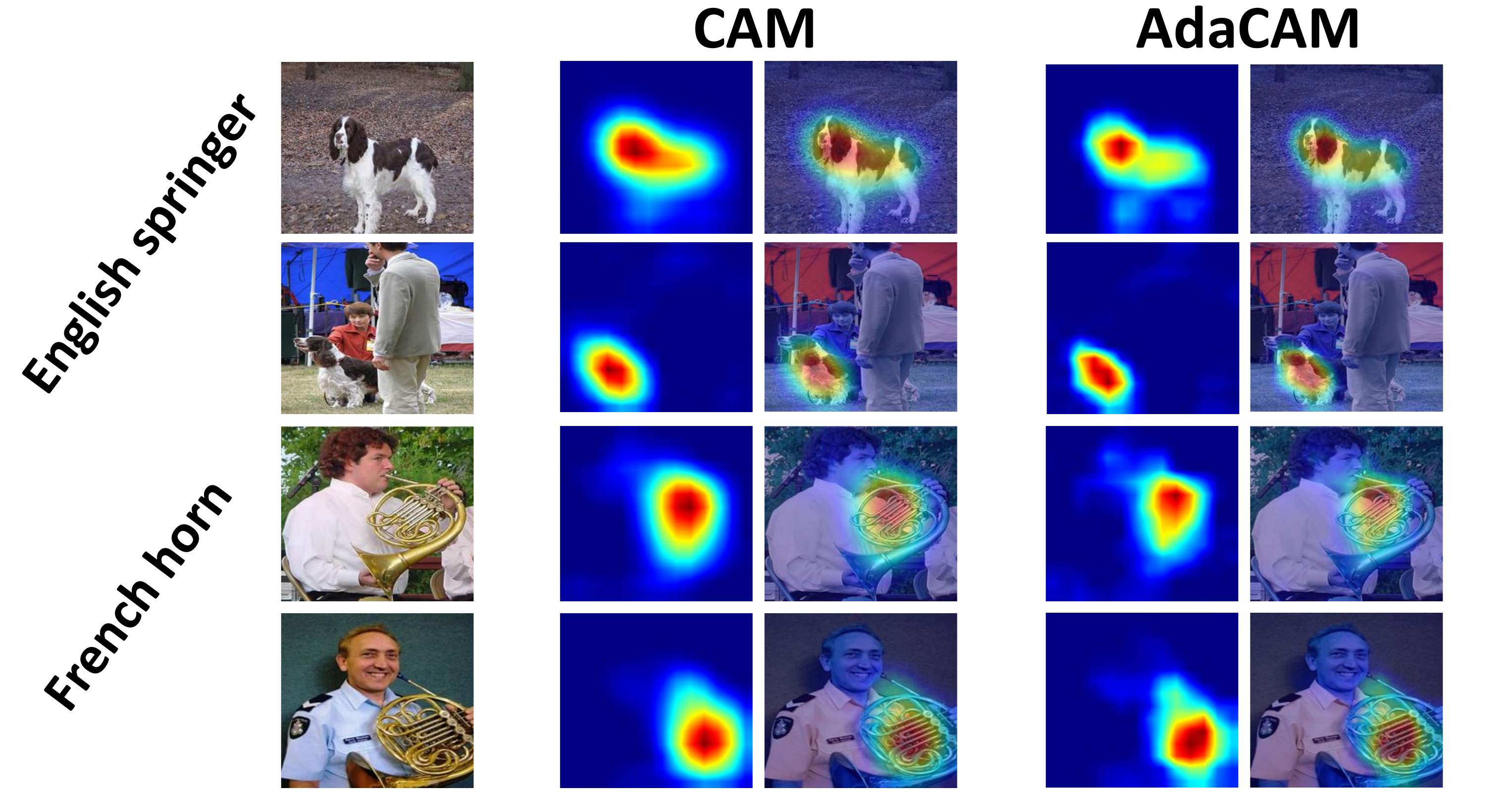}
\caption{Visual comparison between CAM and our AdaCAM (evaluated on Imagenette validation set) in object localization. Refer to supplementary materials for more results.}
\label{CAM_comparison}
\end{figure}

Firstly, we visually demonstrate that our proposed AdaCAM possesses comparable object localization ability as CAM. To this end, we train a network with a structure of [VGG16 backbone$\rightarrow$GAP$\rightarrow$single FC layer] to generate CAM; and train another network of [VGG16 backbone$\rightarrow$single Conv$_{1\times1}$ layer$\rightarrow$GAP] to synthesize AdaCAM. As shown in Figure \ref{CAM_comparison}, though generated efficiently in a feed-forward manner and independently of class label, the attention maps obtained by our AdaCAM are very close to that of CAM.

\renewcommand{\arraystretch}{1.19}
\begin{table*}[b]  
  \centering  
  \fontsize{8.5}{9}\selectfont  
    \begin{tabular}{|p{24mm}<{\centering}||p{12mm}<{\centering}p{12mm}<{\centering}p{10mm}<{\centering}p{10mm}<{\centering}p{10mm}<{\centering}p{10mm}<{\centering}p{10mm}<{\centering}p{10mm}<{\centering}p{10mm}<{\centering}p{10mm}<{\centering}|}
    % \toprule[1pt]
    \hline  
    \multirow{2}{*}{\textbf{Model}}&  
    \multicolumn{10}{c|}{\textbf{Datasets}}\cr\cline{2-11}  
    &Imagenette&Caltech101&Corel5k&Scene15&Indoor67&Action40&Event8&UCMLU&RSSCN7&AID\cr  
    \hline
    % \midrule[1.2pt]
    \hline  
    VGG16\_GAP&82.04\%&70.42\%&54.31\%&80.86\%&68.25\%&52.19\%&80.00\%&90.95\%&91.25\%&87.61\%\cr
    VGG16\_MFA&\textbf{88.45\%}&\textbf{76.05\%}&\textbf{70.74\%}&\textbf{85.53\%}&\textbf{75.27\%}&\textbf{60.36\%}&\textbf{85.75\%}&\textbf{96.67\%}&\textbf{95.29\%}&\textbf{94.55\%}\cr\hline
    % \midrule[1.2pt]
    \hline  
    ResNet50\_GAP&83.66\%&71.84\%&54.55\%&81.41\%&70.14\%&53.43\%&81.08\%&91.67\%&91.43\%&89.48\%\cr  
    ResNet41\_GAP&82.47\%&70.75\%&52.72\%&80.17\%&68.90\%&52.56\%&79.88\%&90.58\%&90.54\%&88.31\%\cr  
    ResNet41\_MFA&\textbf{88.60\%}&\textbf{76.19\%}&\textbf{69.84\%}&\textbf{85.00\%}&\textbf{75.59\%}&\textbf{60.61\%}&\textbf{85.25\%}&\textbf{96.43\%}&\textbf{94.71\%}&\textbf{94.89\%}\cr\hline 
    % \midrule[1.2pt]
    \hline 
    ResNeXt50\_GAP&84.10\%&72.65\%&56.20\%&83.18\%&71.26\%&54.71\%&82.95\%&92.65\%&92.32\%&90.24\%\cr  
    ResNeXt41\_GAP&82.79\%&71.54\%&55.66\%&81.84\%&69.95\%&53.50\%&81.89\%&91.79\%&91.18\%&88.96\%\cr
    ResNeXt41\_MFA&\textbf{88.86\%}&\textbf{76.78\%}&\textbf{71.48\%}&\textbf{86.29\%}&\textbf{76.38\%}&\textbf{61.77\%}&\textbf{87.14\%}&\textbf{97.31\%}&\textbf{95.14\%}&\textbf{95.29\%}\cr\hline
    % \midrule[1.2pt]
    \hline  
    MobileNet\_GAP&81.26\%&70.26\%&53.86\%&80.24\%&67.79\%&52.48\%&80.50\%&90.24\%&90.54\%&86.87\%\cr  
    MobileNet24\_GAP&80.60\%&69.78\%&51.77\%&79.34\%&66.96\%&51.73\%&79.25\%&89.37\%&89.61\%&86.04\%\cr
    MobileNet24\_MFA&\textbf{87.55\%}&\textbf{75.74\%}&\textbf{69.31\%}&\textbf{84.97\%}&\textbf{74.53\%}&\textbf{59.95\%}&\textbf{84.75\%}&\textbf{95.48\%}&\textbf{93.82\%}&\textbf{93.52\%}\cr\hline
    % \bottomrule[1pt]
    \end{tabular}  
	\caption{Comparison between GAP and our MV-FeaAug (MFA) module with $K=50$ in validation accuracy based on different CNN backbones. Our MV-FeaAug impressively boosts model performance with simply one more $Conv_{1\times1}$ layer than GAP-based counterpart.}
\label{tab:performance_comparison} 
\end{table*}

We incorporate our MV-FeaAug module to different CNN backbones as a substitute of GAP and achieve consistent and noticeable performance gains. The chosen CNN backbones are VGG16 \cite{vggnet}, ResNet50 \cite{resnet}, ResNeXt50 \cite{ResNeXt}, and MobileNet \cite{mobilenets}. For 224$\times$224 input images, the output of VGG16 backbone is 14$\times$14 feature maps, upon which we respectively apply GAP and our MV-FeaAug (MFA) module, yielding VGG16\_GAP and VGG16\_MFA respectively. For ResNet50, ResNeXt50, and MobileNet backbones that contain 5 convolutional stages, the output feature maps are of 7$\times$7 spatial size. We firstly build GAP-based models by applying GAP over them, leading to ResNet50\_GAP, ResNeXt50\_GAP, and MobileNet\_GAP. For evaluation of MV-FeaAug, since 7$\times$7 resolution is too small to sample sufficient multi-view local features, we remove the last convolutional stage of these backbones, yielding the truncated counterparts ResNet41, ResNeXt41, and MobileNet24 that have 14$\times$14 output feature maps, upon which we apply MV-FeaAug module to build ResNet41\_MFA, ResNeXt41\_MFA, and MobileNet24\_MFA respectively. For fair comparison between GAP and MV-FeaAug under the same backbones, we also build ResNet41\_GAP, ResNeXt41\_GAP, and MobileNet24\_GAP by applying GAP over the truncated backbones. We set $K=50$ and $L_{R}=[3,5,7,9]$ in our MV-FeaAug module. The used classifier head of all the built models in this section is a simple single-FC-layer softmax classifier, though it is demonstrated in supplementary materials that our module also brings impressive performance gains in the case of more complex classifier head.

\begin{figure}[t]
\includegraphics[width=3.2in]{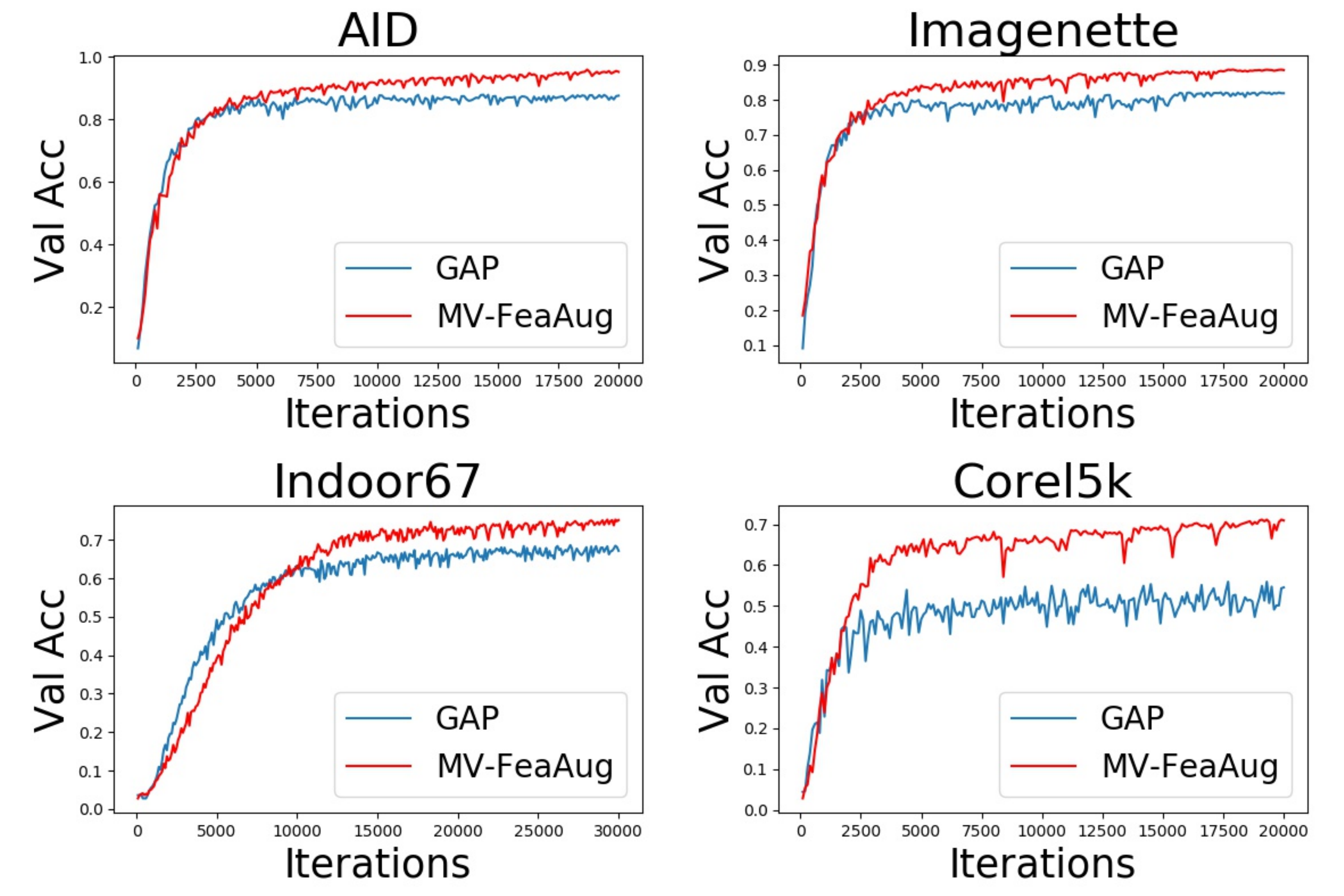}
\caption{Learning curves comparison between GAP and MV-FeaAug on different datasets. A validation accuracy (Val Acc) is evaluated after every 100 training iterations. The used CNN backbone is VGG16. Refer to supplementary materials for more results.}
\label{learning_curves}
\end{figure}

We train 20k iterations on Imagenette, Caltech101, AID, and Corel5k, 10k iterations on Scene15, Event8, and RSSCN7, 15k iterations on UCMLU, 30k iterations on Action40, and 40k iterations on Indoor67, making sure that learning converges on all datasets. The validation accuracies achieved by different models on different datasets are reported in Table \ref{tab:performance_comparison}. It is shown that under the same CNN backbone, image classification performance could be improved by a noticeable margin by substituting our MV-FeaAug module for GAP. Furthermore, we demonstrate that GAP based models with a deeper CNN backbone could be surpassed by our MV-FeaAug based models with a truncated shallower backbone, which shows potential value of our method in model compression. Learning curves of VGG16\_GAP and VGG16\_MFA evaluated on the validation set of different datasets are shown in Figure \ref{learning_curves}, our MV-FeaAug module boosts performance consistently on different datasets.

\renewcommand{\arraystretch}{1.11}

\begin{table}[t]  
  \centering  
  \fontsize{8}{9.5}\selectfont   
    \begin{tabular}{|p{24mm}<{\centering}|p{9mm}<{\centering}p{7mm}<{\centering}p{9mm}<{\centering}|p{12mm}<{\centering}|}
    % \toprule[1.5pt]
    \hline  
    \multirow{2}{*}{\textbf{Model}}&  
    \multicolumn{3}{c|}{\textbf{Datasets}}&
    \multirow{2}{*}{\makecell{\textbf{Backbone} \\ \textbf{Parameters}}}\cr\cline{2-4} 
    &Caltech101&AID&Indoor67&\cr\hline
    % \midrule[1.2pt]
    \hline  
    VGG16\_GAP&70.42\%&87.61\%&68.25\%&14.03M\cr
    % \midrule[1.2pt] 
    \hline
    VGG16\_SE\_GAP&70.96\%&88.44\%&68.75\%&14.18M\cr
	VGG16\_CBAM\_GAP&71.53\%&89.37\%&69.24\%&14.61M\cr  
    VGG16\_SK\_GAP&71.25\%&89.63\%&69.08\%&19.55M\cr
    VGG16\_GC\_GAP&71.87\%&90.04\%&69.51\%&14.76M\cr
    % \midrule[1.2pt]
    \hline 
    VGG16\_mixup\_GAP&71.55\%&89.50\%&69.67\%&14.03M\cr
    VGG16\_cutout\_GAP&70.90\%&88.25\%&69.14\%&14.03M\cr
    VGG16\_cutmix\_GAP&72.10\%&90.16\%&70.03\%&14.03M\cr
    % \midrule[1.2pt]
    \hline  
    VGG16\_MFA (ours)&\textbf{76.05}\%&\textbf{94.55}\%&\textbf{75.27}\%&14.03M\cr\hline  
    % \bottomrule[1.5pt]
    \end{tabular}
    \caption{Comparison of our MV-FeaAug (MFA) with related visual attention modules and data augmentation methods in validation accuracy. We use VGG16 backbone for all the related methods, and set $K=50$, $L_{R}=[3,5,7,9]$ in our MV-FeaAug module.} 
    \label{tab:method_comparison}   
\end{table} 

\renewcommand{\arraystretch}{1.15}
\begin{table}[t]  
  \centering  
  \fontsize{8}{9.5}\selectfont    
    \begin{tabular}{|p{12mm}<{\centering}|p{7mm}<{\centering}|p{7mm}<{\centering}p{7mm}<{\centering}p{7mm}<{\centering}p{7mm}<{\centering}p{7mm}<{\centering}|}
    % \toprule[1.5pt]
    \hline  
    \multirow{2}{*}{\textbf{Datasets}}& 
    \multirow{2}{*}{\textbf{GAP}}& 
    \multicolumn{5}{c|}{\textbf{MV-FeaAug}}\cr\cline{3-7}&
    &K=20&K=30&K=40&K=50&K=60\cr\hline
    % \midrule[1.2pt]
    \hline  
    Imagenette&82.04\%&86.27\%&87.62\%&88.13\%&88.45\%&88.72\%\cr\hline 
    Indoor67&68.25\%&73.51\%&74.34\%&74.89\%&75.27\%&75.55\%\cr\hline 
    Action40&53.21\%&58.24\%&59.36\%&59.98\%&60.36\%&60.68\%\cr\hline 
    UCMLU&90.95\%&94.52\%&95.33\%&96.02\%&96.67\%&97.05\%\cr\hline 
    \end{tabular} 
    % \caption{Validation accuracy on different datasets achieved by MV-FeaAug with different $K$ value. The backbone network is VGG16.} 
    \caption{Study of the impact of $K$ to validation accuracy on different datasets. The backbone network is VGG16.}
    \label{tab:impact_of_K} 
\end{table} 

Since our method relates to both attention mechanism and data augmentation, we quantitatively compare our MV-FeaAug with methods of both two fields. Based on VGG16 backbone, we separately attach a SE\_block \cite{SENet}, a CBAM\_block \cite{cbam}, a GC\_block \cite{gcnet} to the end of each convolutional stage of VGG16 backbone, introducing channel-wise attention, channel-wise and spatial attention, global context self-attention to the network respectively, yielding VGG16\_SE, VGG16\_CBAM, and VGG16\_GC backbones. Besides, we replace the last convolutional layer of each convolutional stage of VGG16 backbone with a SK\_block \cite{SKNet} that combines channel-wise attention with dynamic kernel-size selection, yielding VGG16\_SK backbone. For SE\_block, CBAM\_block, and SK\_block, we set the reduction ratio $r=2$. As shown in Table \ref{tab:method_comparison}, though injecting these attention modules can improve model performance as compared to the baseline model VGG16\_GAP, the effects are very limited compared with our MV-FeaAug module (i.e., VGG16\_MFA), which could be due to that our method (\RNum{1}) explicitly attends to class-related local features; (\RNum{2}) augments training samples at feature space by sampling diverse multi-view features, while the compared attention modules do not possess such object localization and data augmentation abilities. Moreover, adding these attention modules increase backbone network parameters, while our method improves performance remarkably without enlarging backbone network. In addition, as Table \ref{tab:method_comparison} shows, our method also noticeably surpasses classic image-level data augmentation methods, e.g., Mixup \cite{mixup}, Cutout \cite{cutout}, and CutMix \cite{cutmix} (we set $\alpha_{mixup}$=0.2 and mask size to be 56$\times$56 in these methods), since the augmented multi-view features in our module collaboratively contribute to the final prediction of the image, while these methods do not involve such ensemble learning functionality. Ablation study of the contributing factors of our MV-FeaAug module to classification performance is demonstrated in supplementary materials.

\begin{figure}[t]
\includegraphics[width=3.3in]{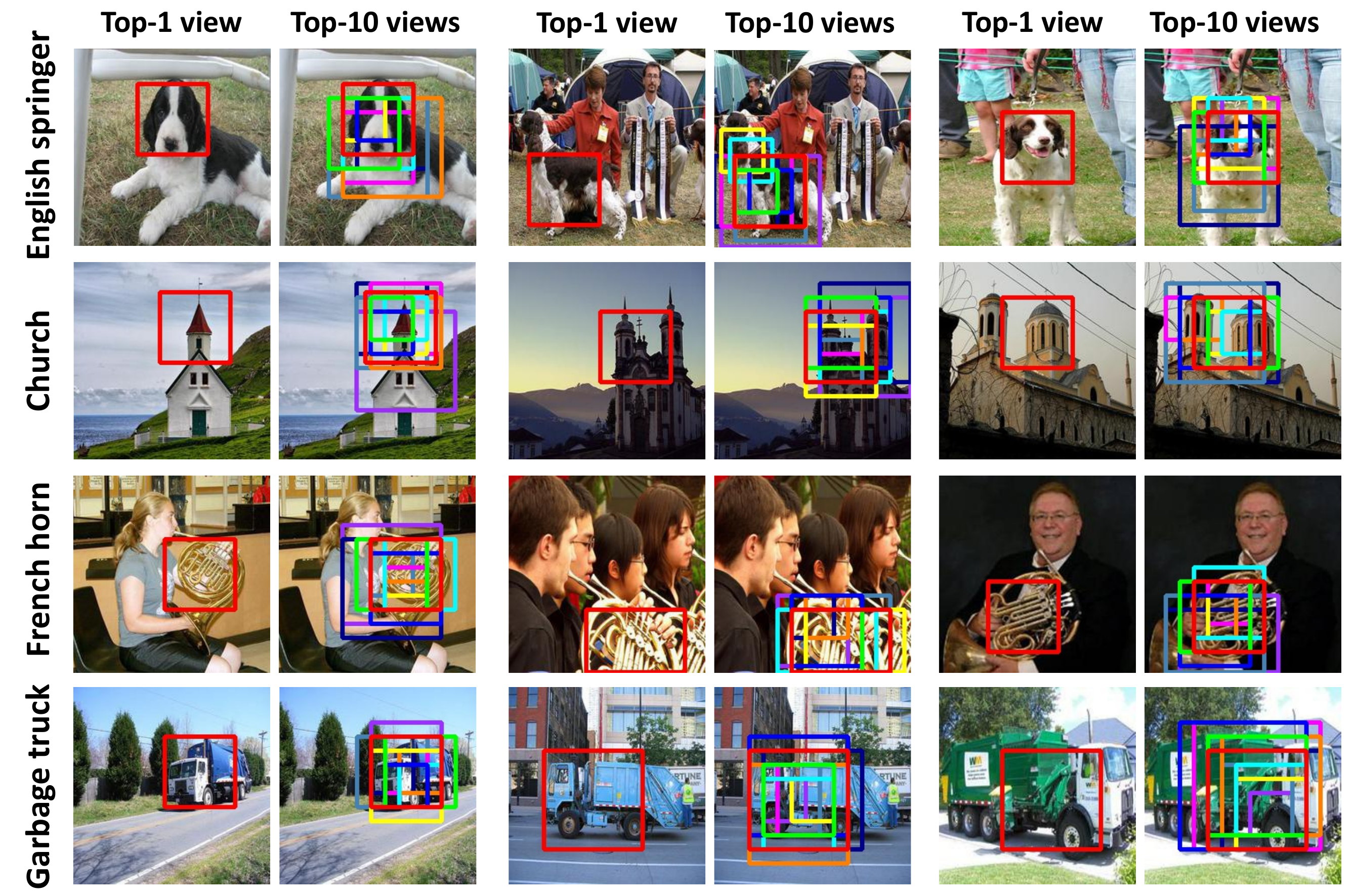}
\caption{Visualization of the top-1 and top-10 local regions that correspond to local features with the highest prediction confidence of the corresponding image class. Please zoom in for better resolution and refer to supplementary materials for more results.}
\label{top_k_views}
\end{figure}

We also explore the influence of $K$ to classification performance based on VGG16\_MFA model. Results in Table \ref{tab:impact_of_K} shows that validation accuracy improves with the increase of $K$, indicating that sufficient number of the augmented multi-view features is important to model performance. 

Based on VGG16\_MFA, we select from all the sampled $K\times R$ multi-view local regions top-k ones that correspond to local features with the highest prediction confidence of the corresponding image class in the main classifier head, and map these top-k regions from feature scale to image scale as visualization of our MV-FeaAug. Results of top-1 and top-10 local regions evaluated on images of the Imagenette validation set are shown in Figure \ref{top_k_views}, we see that the augmented multi-view local features correspond to the selected local regions that basically gather around the image-class-related objects, which demonstrates the effectiveness of our method in extracting rich class-related local features as a kind of attention-based feature-level data augmentation.

\section{Conclusion}
This paper proposes MV-FeaAug, a plug-and-play module performing feature augmentation by sampling multi-view class-related local features with our proposed AdaCAM. By virtue of feature augmentation and multi-view ensemble learning, MV-FeaAug boosts image classification noticeably.

\appendix

\section{Datasets}
We evaluate our method on the following object, scene, and action classification datasets: Caltech101 \cite{caltech101}, Imagenette \cite{imagenette}, Corel5k \cite{corel5k}, Scene15 \cite{scene15}, MIT Indoor67 \cite{indoor67}, Stanford Action40 \cite{action40}, UIUC Event8 \cite{event8}, UCMLU \cite{UCMerced}, RSSCN7 \cite{RSSCN7}, and AID \cite{AID}. 
\begin{itemize}
\item \textbf{Caltech101} consists of pictures of objects belonging to 101 classes. Each class contains roughly 40 to 800 images, totalling 8679 images. The training and validation set contain 6983 and 1696 images respectively.
\item \textbf{Imagenette} is a subset of 10 classes from the Imagenet dataset with totally 13396 images. The training and validation set contain 9470 and 3926 images respectively.
\item \textbf{Corel5k} is a dataset containing 5000 images that is normally used for image classification and retrieval. It contains 50 categories with 100 images per class. We use 4500 images for training and the rest 500 images for validation. Both training and validation set have equal number of images per class. 
\item \textbf{Scene15} is a 15 class scene classification dataset comprising office, kitchen, living room, bedroom, store, industrial, tall building, inside cite, street, highway, coast, open country, mountain, forest, and suburb. It contains totally 4486 images from which we split the training and validation set with a proportion of 3:1.
\item \textbf{Indoor67} is a 67 class indoor scene (e.g., bookstore, bar, casino, corridor, laundromat, art painting studio, etc.) classification dataset with a total of 15620 images. We split training and validation set with a proportion of 3:1.
\item \textbf{Stanford Action40} is a dataset containing images of humans performing 40 actions (e.g., applauding, brushing teeth, cleaning the floor, climbing, drinking, smoking, etc.). It has totally 9532 images from which we split training and validation set with a proportion of 10:3.
\item \textbf{UIUC Event8} is a dataset containing 8 sports event categories: rowing, badminton, polo, bocce, snowboarding, croquet, sailing, and rock climbing, with totally 1587 images. We split 1100 images for training and the rest 487 images for validation.
\item \textbf{UCMLU} (UC Merced Land Use) is a 21 class land use remote sensing (e.g., buildings, harbor, beach, tenniscourt, etc.) image dataset with 100 images per class. The training set contains 80 images per class and the validation set contains 20 images per class.
\item \textbf{RSSCN7} is remote sensing dataset containing 7 classic scene categories: grass, forest, farmland, parking lot, residential, industrial, and laker. It has 400 images per class and 2800 images in total. We split a training set with 320 images per class and a validation set with 80 images per class.
\item \textbf{AID} is a large-scale aerial image dataset made up of 30 aerial scene types, e.g., airport, bare land, beach, commercial, meadow, playground, stadium, etc. It contains totally 10000 images from which we split 8225 images for training and the rest 1775 images for validation.
\end{itemize}

\begin{figure*}[t]
\center
\includegraphics[width=6.6in]{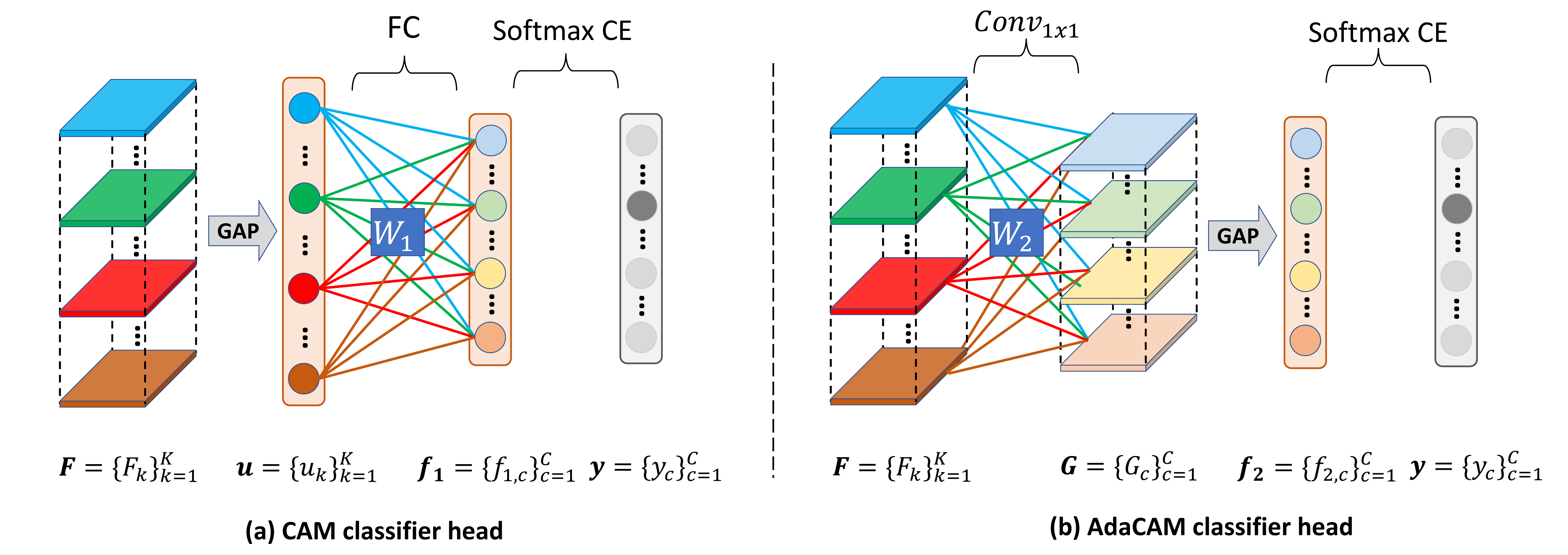}
\caption{Comparison between (a) the classifier head of CAM and (b) the modified classifier head of AdaCAM.}
\label{CAM_vs_AdaCAM}
\end{figure*}

\section{Proof: AdaCAM is essentially an efficient approximation to CAM}
Comparison between the classifier head for CAM and our modified classifier head for AdaCAM is illustrated in Figure \ref{CAM_vs_AdaCAM}. Let $\textit{\textbf{F}}=\{F_{k}\}_{k=1}^{K}$ be the output convolutional feature maps of the CNN backbone, the feature maps $\textit{\textbf{F}}$ has $K$ channels. For the classifier head of CAM, the GAP is firstly applied to vectorize feature maps $\textit{\textbf{F}}$ into a vector $\textbf{\textit{u}}=\{u_{k}\}_{k=1}^{K}$ with $K$ units. Then a fully-connected layer is used to map $\textbf{\textit{u}}$ into the logit vector $\textbf{\textit{f}}_{1}=\{f_{1,c}\}_{c=1}^{C}$ that has the same number of $C$ units as the number of classes. The loss function is the softmax cross-entropy loss between the logit vector $\textbf{\textit{f}}_{1}$ and the one-hot label vector $\textbf{\textit{y}}=\{y_{c}\}_{c=1}^{C}$. For classifier head of our AdaCAM, we firstly use a $1\times1$ convolutional layer to convert feature maps $\textit{\textbf{F}}$ to feature maps $\textbf{\textit{G}}=\{G_{c}\}_{c=1}^{C}$ that have the same number of $C$ channels as the number of classes. Then we use GAP to vectorize feature maps $\textit{\textbf{G}}$ into the logit vector $\textbf{\textit{f}}_{2}=\{f_{2,c}\}_{c=1}^{C}$, the loss function is the softmax cross-entropy loss between the logit vector $\textbf{\textit{f}}_{2}$ and the one-hot label vector $\textbf{\textit{y}}$. Below we prove that the attention map of our AdaCAM is approximately the same as that of CAM.

\proof Refer to Figure \ref{CAM_vs_AdaCAM}, starting from the CNN backbone output feature maps $\textit{\textbf{F}}=\{F_{k}\}_{k=1}^{K}$, the CAM classifier head obtains the final logit vector $\textbf{\textit{f}}_{1}$ with GAP followed by a FC layer with weight matrix $\textbf{\textit{W}}_{1} \in R^{K\times C}$:

\begin{align}
	\textbf{\textit{u}}=[u_{1},u_{2},...,u_{K}]^{T},
\end{align}
\begin{align}
	u_{k}=\frac{1}{HW}\sum\nolimits_{x,y}F_{k}(x,y),
\end{align}
\begin{align}
	\textbf{\textit{f}}_{1}=\textbf{\textit{W}}_{1}^{T}\textbf{\textit{u}},
\end{align}
where each column of the weight matrix $\textbf{\textit{W}}_{1}$ is a weight vector corresponding to each unit of the logit vector $\textbf{\textit{f}}_{1}$:

\begin{align}
	\textbf{\textit{W}}_{1}=[\textbf{\textit{w}}_{1,1},\textbf{\textit{w}}_{1,2},...,\textbf{\textit{w}}_{1,C}], 
\end{align}
\begin{align}
	\textit{f}_{1,i}=\textbf{\textit{w}}_{1,i}^{T}\textbf{\textit{u}}, \quad i=1,2,...,C,
\end{align}
where $\textit{f}_{1,i}$ is the $i$-th element of $\textbf{\textit{f}}_{1}$. Substituting equation (2) into equation (5), we obtain that:
\begin{align}
\textit{f}_{1,i}&=\sum\nolimits_{j=1}^{K}\textit{w}_{1,i,j}\textit{u}_{j}\\
				&=\sum\nolimits_{j=1}^{K}\textit{w}_{1,i,j}\frac{1}{HW}\sum\nolimits_{x,y}F_{j}(x,y)\\
				&=\frac{1}{HW}\sum\nolimits_{x,y}\sum\nolimits_{j=1}^{K}\textit{w}_{1,i,j}F_{j}(x,y), \quad i=1,2,...,C,
\end{align}
in which $\textit{w}_{1,i,j}$ is the $j$-th element of the weight vector $\textbf{\textit{w}}_{1,i}$.

For our AdaCAM classifier head, a $Conv_{1\times1}$ layer is firstly applied to map feature maps $\textbf{\textit{F}}=\{F_{k}\}_{k=1}^{K}$ to $\textbf{\textit{G}}=\{G_{c}\}_{c=1}^{C}$. Since $1\times1$ convolution is equivalent to fully-connected layer where each feature map (channel) could be regarded as a unit in a FC layer, it essentially performs weighted combination of feature maps with a weight matrix $\textit{\textbf{W}}_{2} \in R^{K\times C}$:
\begin{align}
	\textbf{\textit{W}}_{2}=[\textbf{\textit{w}}_{2,1},\textbf{\textit{w}}_{2,2},...,\textbf{\textit{w}}_{2,C}], 
\end{align}
\begin{align}
  	G_{i}=\sum\nolimits_{j=1}^{K}F_{j}\textit{w}_{2,i,j}, \quad i=1,2,...,C,
\end{align}
in which $G_{i}$ and $F_{j}$ are the $i$-th channel of $\textbf{\textit{G}}$ and the $j$-th channel of $\textbf{\textit{F}}$ respectively, and $\textit{w}_{2,i,j}$ denotes the $j$-th element of the weight vector $\textbf{\textit{w}}_{2,i}$ that corresponds to $G_{i}$. Since the feature maps $\textbf{\textit{F}}$ and $\textbf{\textit{G}}$ shares the same spatial size, we have:
\begin{align}
  	G_{i}(x,y)=\sum\nolimits_{j=1}^{K}\textit{w}_{2,i,j}F_{j}(x,y), \quad i=1,2,...,C.
\end{align}

After $1\times1$ convolution, the GAP is applied to vectorize feature maps $\textbf{\textit{G}}$ into the logit vector $\textbf{\textit{f}}_{2}$:
\begin{align}
\textit{f}_{2,i}=\frac{1}{HW}\sum\nolimits_{x,y}G_{i}(x,y), \quad i=1,2,...,C,
\end{align}
where $\textit{f}_{2,i}$ is the $i$-th element of $\textbf{\textit{f}}_{2}$.
Substituting equation (11) into equation (12), we obtain that:
\begin{align}
\textit{f}_{2,i}=\frac{1}{HW}\sum\nolimits_{x,y}\sum\nolimits_{j=1}^{K}\textit{w}_{2,i,j}F_{j}(x,y), \quad i=1,2,...,C.
\end{align}

Comparing equation (8) and equation (13), we obtain that $\textbf{\textit{f}}_{1}=\textbf{\textit{f}}_{2}$ if $\textbf{\textit{W}}_{1}=\textbf{\textit{W}}_{2}$. That is, if the FC layer in CAM classifier head and the $Conv_{1\times1}$ layer in AdaCAM classifier head share the same parameters, both two structures result in the same logit vector. Therefore, we conversely obtain that if $\textbf{\textit{f}}_{1}=\textbf{\textit{f}}_{2}$, then $\textbf{\textit{W}}_{1}=\textbf{\textit{W}}_{2}$. 

Since both two logit vectors $\textbf{\textit{f}}_{1}$ and $\textbf{\textit{f}}_{2}$ are driven towards the ground truth one-hot label vector $\textbf{\textit{y}}$ under the minimization of softmax cross-entropy loss, suppose that after sufficient training time, both $\textbf{\textit{f}}_{1}$ and $\textbf{\textit{f}}_{2}$ approximate $\textbf{\textit{y}}$, then we have $\textbf{\textit{f}}_{1} \approx \textbf{\textit{f}}_{2}$, and thus $\textbf{\textit{W}}_{1} \approx \textbf{\textit{W}}_{2}$. At this point, given the ground truth class label $c'$, the attention map obtained by CAM is:
\begin{align}
A_{c'}&=\sum\nolimits_{j=1}^{K}{\textit{w}_{1,c',j}}F_{j}\\
	  &\approx \sum\nolimits_{j=1}^{K}{\textit{w}_{2,c',j}F_{j}}\\
	  &=G_{c'}.
\end{align}
That is, the attention map of CAM approximately equals the $c'$-th channel of $\textbf{\textit{G}}$. For ground truth label $c'$, the target label vector $\textbf{\textit{y}}$ is a one-hot vector with its $c'$-th element being 1 and 0 otherwise, thus we have:
\begin{align}
A_{c'} \approx G_{c'}&=\sum\nolimits_{j=1}^{C}G_{j}\textit{y}_{j}.
\end{align}
On the other hand, the one-hot label vector $\textbf{\textit{y}}$ could be approximated by the softmax activation of the logit vector $\textbf{\textit{f}}_{2}$ after sufficient training steps, since softmax operation exponentially widens the gap between elements of $\textbf{\textit{f}}_{2}$:
\begin{align}
A_{c'}&\approx \sum\nolimits_{j=1}^{C}G_{j}y_{j}\\
      &\approx \sum\nolimits_{j=1}^{C}G_{j}\frac{e^{\textit{f}_{2,j}}}{\sum\nolimits_{k=1}^{C}e^{\textit{f}_{2,k}}}.
\end{align}

That is, the attention map of the target class could be approximated by the weighted sum of feature maps $\{G_{c}\}_{c=1}^{C}$ with respect to the softmax logit vector Softmax($\textbf{\textit{f}}_{2}$), which is exactly the attention map generation process of our AdaCAM. Thus we prove that our AdaCAM is essentially an efficient approximation to CAM. The advantages of our AdaCAM is that (1) it generates attention map with only a feed-forward pass of the network; (2) it is not conditioned on the ground truth label, which allows for adaptive generation of attention map at inference time where the target label is unknown; (3) it is not restricted to a single $Conv_{1\times1}$ layer, but also could use a MLPConv network comprising multiple $Conv_{1\times1}$ layers joined by non-linear activations, as long as the channel number of the last $Conv_{1\times1}$ layer equals to the class number $C$, while CAM is limited to using only a single fully-connected layer.

\begin{figure}[H]
\center
\includegraphics[width=3.3in]{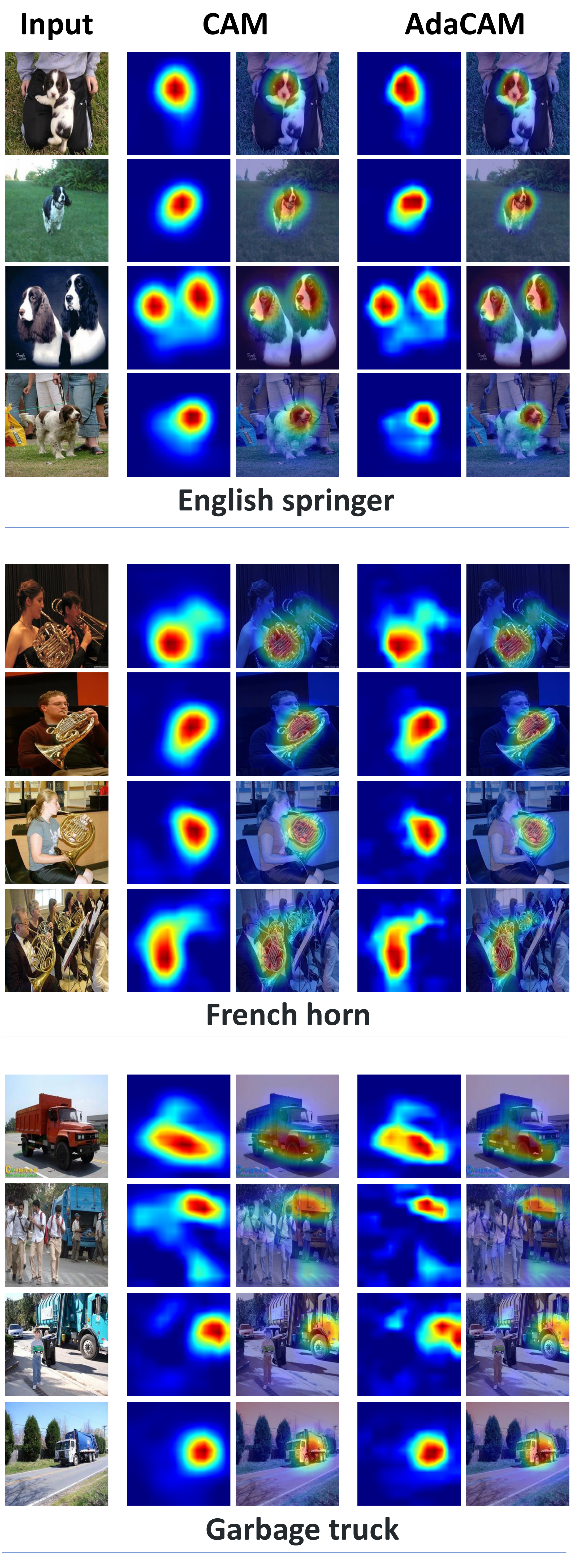}
\caption{Visual comparison between CAM and AdaCAM evaluated on the Imagenette dataset based on VGG16 backbone.}
\label{CAM_vs_AdaCAM_imagenette}
\end{figure}
\begin{figure}[H]
\center
\includegraphics[width=3.3in]{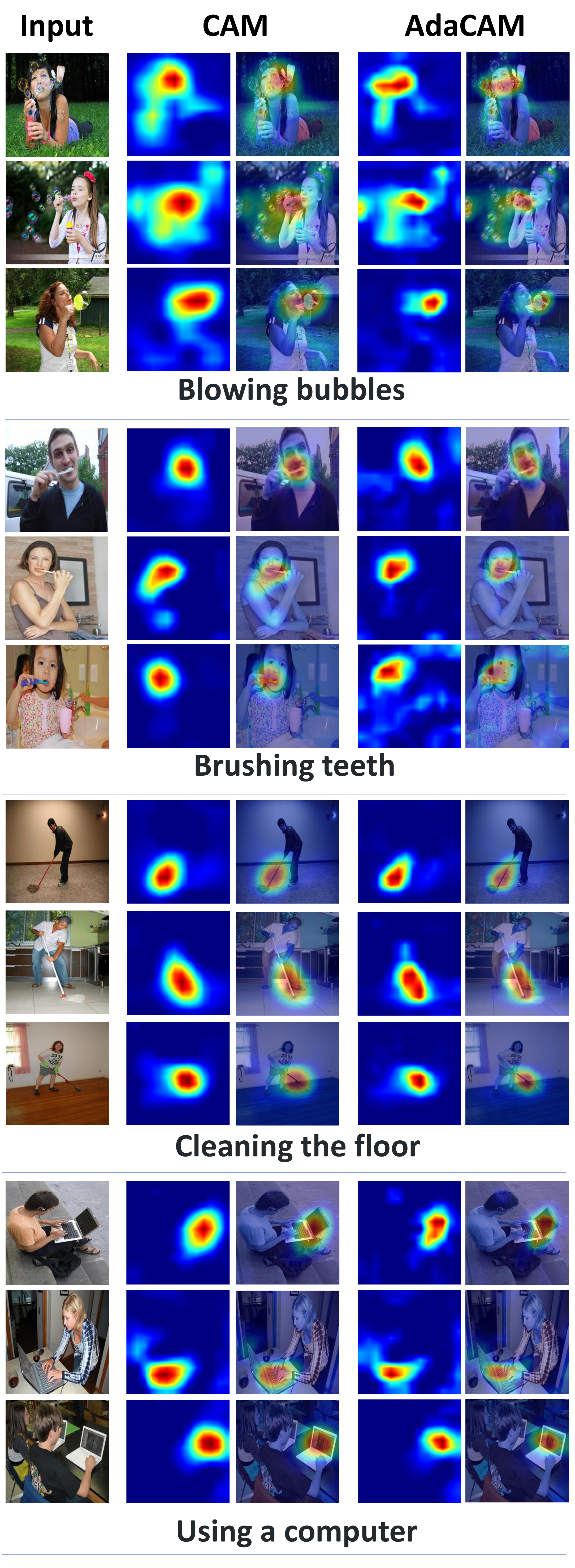}
\caption{Visual comparison between CAM and AdaCAM evaluated on the Stanford Action40 dataset based on VGG16 backbone.}
\label{CAM_vs_AdaCAM_Action40}
\end{figure}
\begin{figure}[H]
\center
\includegraphics[width=3.3in]{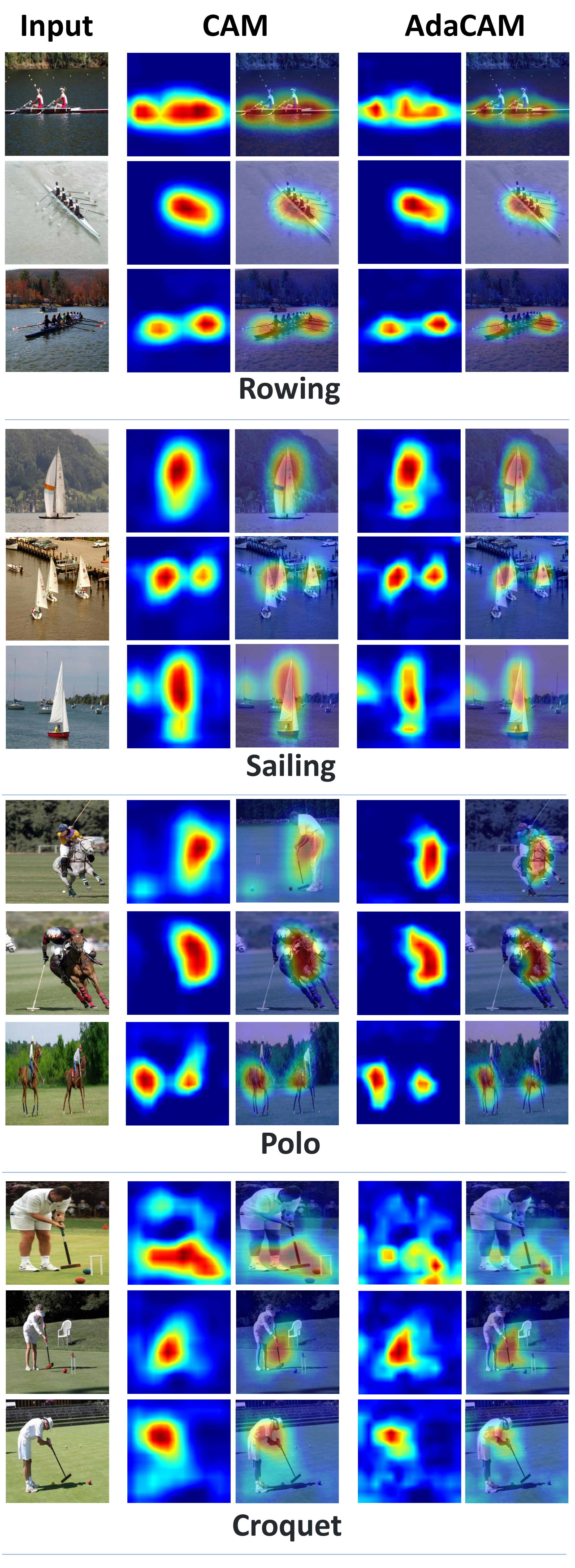}
\caption{Visual comparison between CAM and AdaCAM evaluated on the UIUC Event8 dataset based on VGG16 backbone.}
\label{CAM_vs_AdaCAM_Event8}
\end{figure} 
\section{More visual comparison between CAM and our AdaCAM}
We show more results of visual comparison between CAM and our AdaCAM evaluated on Imagenette, Stanford Action40, and UIUC Event8 dataset in Figure \ref{CAM_vs_AdaCAM_imagenette}, Figure \ref{CAM_vs_AdaCAM_Action40}, and Figure \ref{CAM_vs_AdaCAM_Event8} respectively. Results demonstrate that our AdaCAM achieves similar class-discriminative region localization ability as CAM, despite its advantage of feed-forward and label-independent generation of attention maps. 

\section{More validation accuracy learning curves comparison between GAP and MV-FeaAug}
Below, we display learning curves comparison between GAP and our MV-FeaAug on different datasets. In all the experiments, we use VGG16 backbone and set $K=50$, $L_{R}$ = [3,5,7,9] in our MV-FeaAug module. We evaluate an accuracy rate on the whole validation set after every 100 training iterations. Results show that our MV-FeaAug achieves consistent performance gains compared with GAP.
\begin{figure}[H]
\center
\includegraphics[width=3.3in]{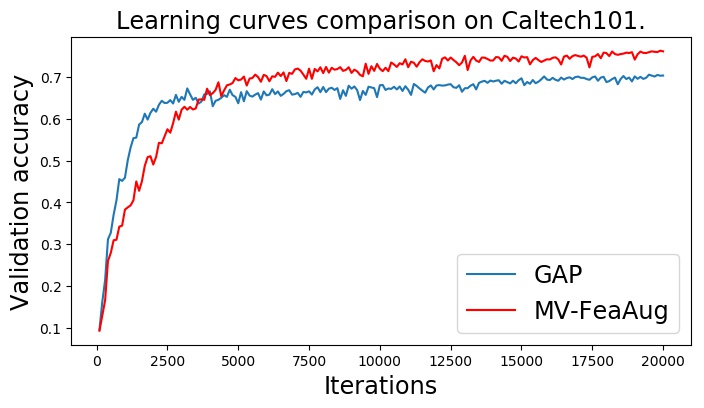}
\caption{Learning curves comparison on Caltech101.}
\label{}
\end{figure} 
\begin{figure}[H]
\center
\includegraphics[width=3.3in]{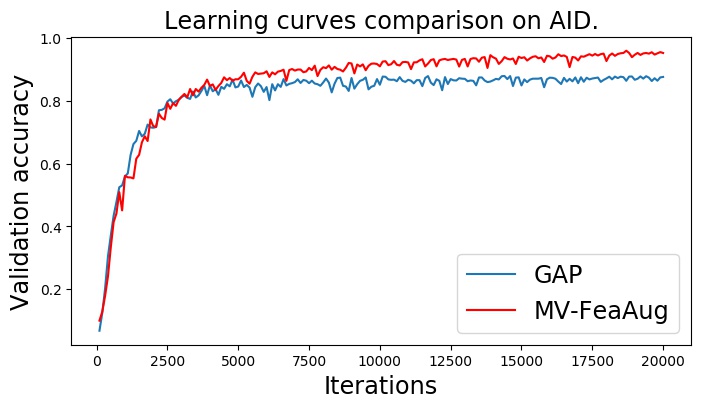}
\caption{Learning curves comparison on AID.}
\label{}
\end{figure} 
\begin{figure}[H]
\center
\includegraphics[width=3.3in]{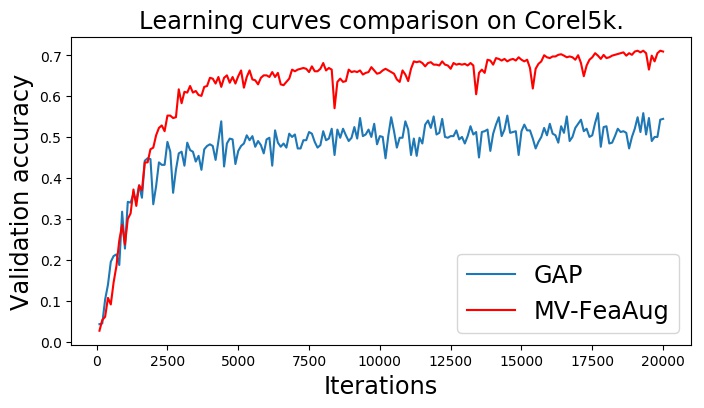}
\caption{Learning curves comparison on Corel5k.}
\label{}
\end{figure} 
\begin{figure}[H]
\center
\includegraphics[width=3.3in]{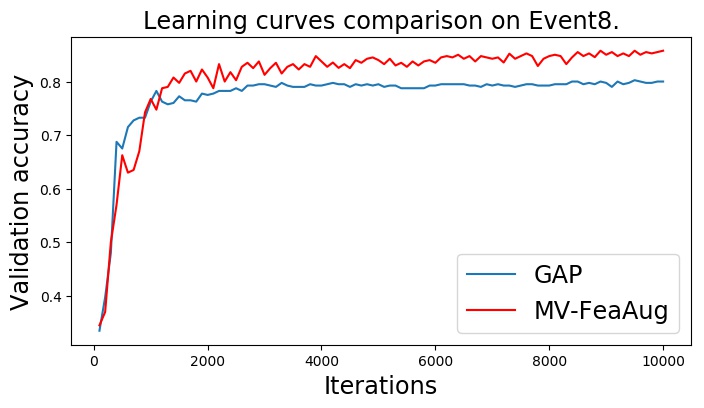}
\caption{Learning curves comparison on UIUC Event8.}
\label{}
\end{figure} 
\begin{figure}[H]
\center
\includegraphics[width=3.3in]{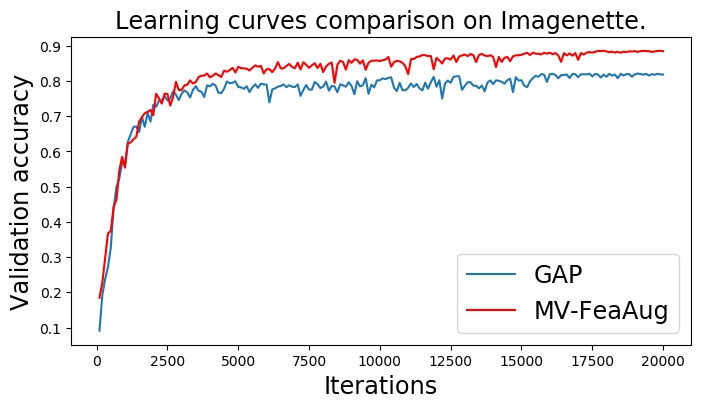}
\caption{Learning curves comparison on Imagenette.}
\label{}
\end{figure} 
\begin{figure}[H]
\center
\includegraphics[width=3.3in]{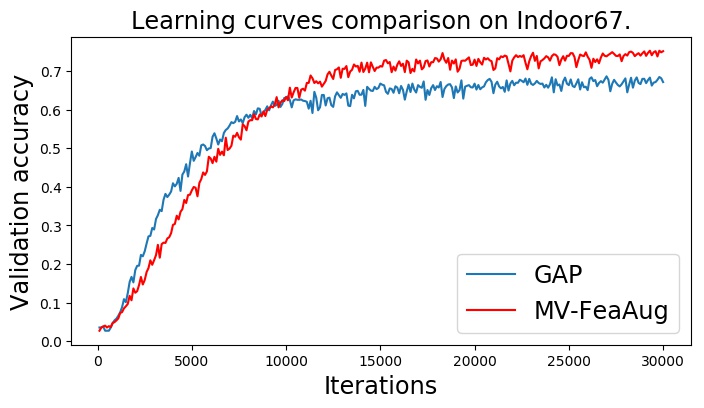}
\caption{Learning curves comparison on MIT Indoor67.}
\label{}
\end{figure} 
\begin{figure}[H]
\center
\includegraphics[width=3.3in]{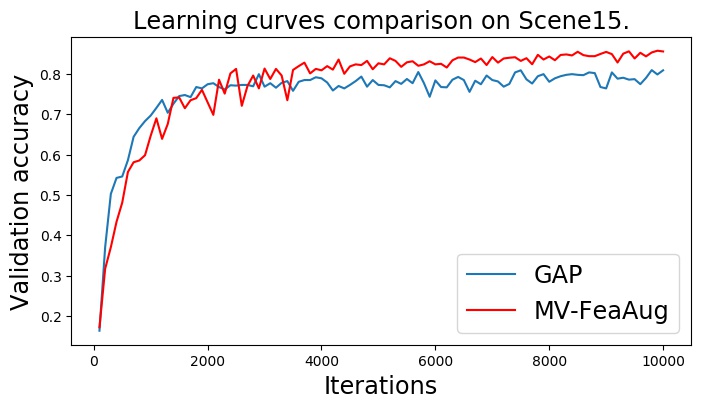}
\caption{Learning curves comparison on Scene15.}
\label{}
\end{figure} 
\begin{figure}[H]
\center
\includegraphics[width=3.3in]{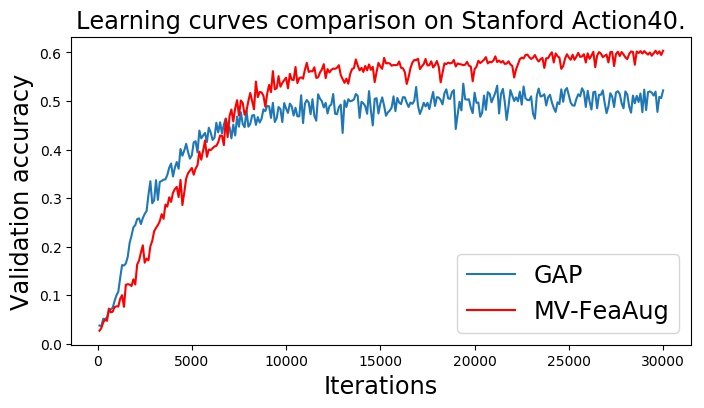}
\caption{Learning curves comparison on Stanford Action40.}
\label{}
\end{figure} 
\begin{figure}[H]
\center
\includegraphics[width=3.3in]{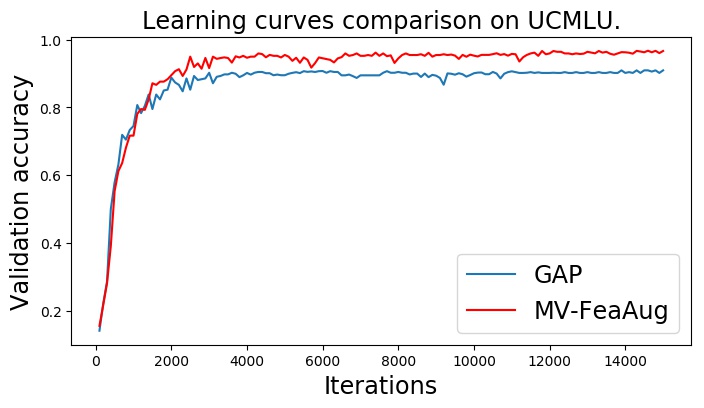}
\caption{Learning curves comparison on UCMLU.}
\label{}
\end{figure}

\section{Experiments with other types of main classifier head}
We previously use a single-FC-layer softmax classifier as the classifier head for both GAP-based and MV-FeaAug-based models, below we investigate the performance gains achieved by our MV-FeaAug module in the case of more complex or more robust classifier head. 
\subsection{MLP classifier head}

Based on VGG16 backbone, we construct several MLP softmax classifier heads to compare our MV-FeaAug with GAP under different classifier heads. The architectures of these classifier heads are described in Table \ref{tab:MLP_classifier_head}.

\renewcommand{\arraystretch}{2}
\begin{table*}[b]  
  \centering  
  \fontsize{8.5}{9.5}\selectfont    
    \begin{tabular}{|l|l|}

    \hline  
    Cls$_{1}$&FC$_{256}\rightarrow$ Relu $\rightarrow$ Dropout $\rightarrow$ FC$_{C}$\cr\hline 
    Cls$_{2}$&FC$_{256}\rightarrow$ Relu $\rightarrow$ Dropout $\rightarrow$ FC$_{128}\rightarrow$ Relu $\rightarrow$ Dropout $\rightarrow$ FC$_{C}$\cr\hline 
    Cls$_{3}$&FC$_{256}\rightarrow$ Relu $\rightarrow$ Dropout $\rightarrow$ FC$_{128}\rightarrow$ Relu $\rightarrow$ Dropout $\rightarrow$ FC$_{128} \rightarrow$ Relu $\rightarrow$ Dropout $\rightarrow$ FC$_{C}$\cr\hline 
    Cls$_{4}$&FC$_{256}\rightarrow$ Relu $\rightarrow$ Dropout $\rightarrow$ FC$_{256}\rightarrow$ Relu $\rightarrow$ Dropout $\rightarrow$ FC$_{128} \rightarrow$ Relu $\rightarrow$ Dropout $\rightarrow$ FC$_{128} \rightarrow$ Relu $\rightarrow$ Dropout $\rightarrow$ FC$_{C}$\cr\hline 
    \end{tabular} 
    \caption{Architecture of the used MLP classifier head Cls$_{1}$, Cls$_{2}$, Cls$_{3}$, and Cls$_{4}$. FC$_{n}$ refers to fully-connected layer with $n$ output units, $C$ denotes the number of classes. The dropout rate for all the dropout layers is 0.5.} 
    \label{tab:MLP_classifier_head} 
\end{table*} 

Validation accuracies on different datasets achieved by GAP and our MV-FeaAug module in combination with different MLP classifier head are reported in Table \ref{tab:performance_with_different_MLP_heads}. We observe that replacing the original single-FC-layer softmax classifier head with MLP ones improves model performance moderately for both GAP and our MV-FeaAug, which could be due to the effect of the intermediate dropout layers in reducing overfitting. In the case of more complex MLP classifier heads, our MV-FeaAug module still boosts model performance by a large margin as compared with GAP.

\renewcommand{\arraystretch}{1.8}
\begin{table}[H]  
  \centering  
  \fontsize{8.5}{9}\selectfont  
    \begin{tabular}{|p{14mm}<{\centering}||p{12mm}<{\centering}p{12mm}<{\centering}p{10mm}<{\centering}p{10mm}<{\centering}|}
    % \toprule[2pt]
    \hline  
    \multirow{2}{*}{\textbf{Model}}&  
    \multicolumn{4}{c|}{\textbf{Datasets}}\cr\cline{2-5}  
    &Imagenette&Indoor67&Action40&AID\cr  
    \hline
    % \midrule[1.2pt]
    \hline  
    Sin\_GAP&82.04\%&68.25\%&52.19\%&87.61\%\cr\hline  
    Sin\_MFA&88.45\%&75.27\%&60.36\%&95.55\%\cr\hline
    \hline  
    Cls$_{1}$\_GAP&82.86\%&68.58\%&53.25\%&88.06\%\cr\hline  
    Cls$_{1}$\_MFA&88.91\%&75.74\%&60.94\%&95.25\%\cr\hline
    % \midrule[1.2pt]
    \hline  
    Cls$_{2}$\_GAP&83.26\%&69.10\%&53.68\%&88.65\%\cr\hline  
    Cls$_{2}$\_MFA&89.18\%&76.08\%&61.40\%&95.58\%\cr\hline  
    % \midrule[1.2pt]
    \hline 
    Cls$_{3}$\_GAP&83.89\%&69.65\%&53.55\%&89.06\%\cr\hline  
    Cls$_{3}$\_MFA&90.03\%&76.42\%&61.29\%&95.93\%\cr\hline
    % \midrule[1.2pt]
    \hline  
    Cls$_{4}$\_GAP&83.50\%&70.06\%&53.84\%&88.76\%\cr\hline  
    Cls$_{4}$\_MFA&89.77\%&76.98\%&61.90\%&95.71\%\cr\hline
    % \bottomrule[2pt]
    \end{tabular} 
	\caption{Comparison between GAP and MV-FeaAug ($K=50$) in validation accuracy based on different MLP classifier heads. The used CNN backbone is VGG16. Cls$_{k}$\_GAP and Cls$_{k}$\_MFA respectively denotes GAP-based and MV-FeaAug-based model where the used classifier head is Cls$_{k}$. The architecture of Cls$_{k}$ ($k$=1,2,3,4) is described in Table \ref{tab:MLP_classifier_head}. Sin\_GAP and Sin\_MFA respectively denotes GAP-based and MV-FeaAug-based model that uses single-FC-layer softmax classifier head.}
\label{tab:performance_with_different_MLP_heads}
\end{table}

\subsection{Prototype classifier head}
Different from discriminative model based softmax classifier head, we also experiment on generative model based prototype classifier head, which learns prototypes for each class and uses prototype matching for classification. For simplicity, we learn only one prototype for each class, since we empirically find that increasing the number of prototypes brings marginal performance gains on our used datasets. Below we formulate the prototype learning framework based on GAP and our MV-FeaAug respectively.

\subsubsection{GAP based prototype classifier}
Starting from the CNN backbone output feature maps $\textbf{\textit{F}}=\{F_{k}\}_{k=1}^{K}$ that have $K$ channels. The normal prototype classifier firstly uses GAP to extract a vector representation $\textbf{\textit{u}}$=GAP($\textbf{\textit{F}}$), then a fully-connected layer is applied to map $\textbf{\textit{u}}$ into a dimension-reduced feature vector $\textbf{\textit{f}}$, i.e., $\textbf{\textit{f}}$=FC($\textbf{\textit{u}}$), where the dimension of $\textbf{\textit{f}}$ is not necessarily to be the same as the number of classes $C$. The prototype classifier head maintains and learns a prototype for each class, which are denoted as $\textbf{\textit{M}}=\{\textbf{\textit{m}}_{j}\}_{j=1}^{C}$, where each prototype $\textbf{\textit{m}}_{j}$ is a randomly initialized learnable vector representation with the same dimension as $\textbf{\textit{f}}$. We use the MCE and DCE loss functions proposed in \cite{prototype} to train the model, which are defined as below.

\noindent\textbf{(1) Minimum classification error loss (MCE)}

\noindent For MCE loss, a misclassification measure $s$ is defined as
\begin{align}
	s=||\textbf{\textit{f}}-\textbf{\textit{m}}_{y}||_{2}^{2}-||\textbf{\textit{f}}-\textbf{\textit{m}}_{r}||_{2}^{2},
\end{align}
in which $\textbf{\textit{m}}_{y}$ is the prototype of genuine class $\textit{y}$, i.e., the correct class corresponding to $\textbf{\textit{f}}$; $\textbf{\textit{m}}_{r}$ denotes the closest prototype from incorrect classes. Then, the MCE loss is defined as:
\begin{align}
	L_{MCE}=\frac{1}{1+e^{-s}}.
\end{align}
The minimization of MCE loss makes the model learn a proper prototype for each class, and drives the extracted vector representation $\textbf{\textit{f}}$ towards the prototype of the correct class and apart from the prototypes of the incorrect classes. The predicted class at inference time is the class whose prototype is closest to vector \textbf{\textit{f}}.

\noindent \textbf{(2) Distance based cross entropy loss (DCE)}

\noindent For DCE loss, the probability of vector representation $\textbf{\textit{f}}$ belonging to the $i$-th class is modelled with the distance between $\textbf{\textit{f}}$ and $\textbf{\textit{m}}_{i}$ (the prototype of the $i$-th class). That is:
\begin{align}
	p(\textbf{\textit{f}} \in C_{i}) \propto -||\textbf{\textit{f}}-\textbf{\textit{m}}_{i}||_{2}^{2}.
\end{align}
Then, softmax operation is utilized to satisfy non-negative and sum-to-one properties of the probability:
\begin{align}
	p(\textbf{\textit{f}} \in C_{i})=\frac{e^{-||\textbf{\textit{f}}-\textbf{\textit{m}}_{i}||_{2}^{2}}}{\sum\nolimits_{j=1}^{C}{e^{-||\textbf{\textit{f}}-\textbf{\textit{m}}_{j}||_{2}^{2}}}}.
\end{align}
Therefore, the DCE loss is defined as:
\begin{align}
	L_{DCE}=-\log\frac{e^{-||\textbf{\textit{f}}-\textbf{\textit{m}}_{y}||_{2}^{2}}}{\sum\nolimits_{j=1}^{C}{e^{-||\textbf{\textit{f}}-\textbf{\textit{m}}_{j}||_{2}^{2}}}},
\end{align}
in which $y$ is the corresponding class of $\textbf{\textit{f}}$, and $\textbf{\textit{m}}_{y}$ is the prototype of class $y$. Similarly, the predicted class at inference time is determined by which prototype is closest to vector \textbf{\textit{f}}.

\subsubsection{MV-FeaAug based prototype classifier}

Below we describe how to combine our MV-FeaAug module with prototype classifier head. For the augmented feature vectors $\{\textbf{\textit{g}}_{i}\}_{i=1}^{K\times R}$ ($K$ and $R$ are the number of anchors per image and the number of multi-scale regions per anchor respectively), a fully-connected layer is utilized to map $\{\textbf{\textit{g}}_{i}\}_{i=1}^{K\times R}$ to dimension-reduced vector representations $\{\textbf{\textit{v}}_{i}\}_{i=1}^{K\times R}$, i.e., $\{\textbf{\textit{v}}_{i}\}_{i=1}^{K\times R}$=FC($\{\textbf{\textit{g}}_{i}\}_{i=1}^{K\times R}$), each $\textbf{\textit{v}}_{i}$ is a local image representation extracted from a specific view. Then, the loss function is the sum of the MCE loss or DCE loss over each representation $\textbf{\textit{v}}_{i},i=1,2,...,K\times R$:
\begin{align}
	L_{MFA\_MCE}=\sum\nolimits_{i=1}^{K\times R}\frac{1}{1+e^{-(||\textbf{\textit{v}}_{i}-\textbf{\textit{m}}_{y}||_{2}^{2}-||\textbf{\textit{v}}_{i}-\textbf{\textit{m}}_{r,i}||_{2}^{2})}},
\end{align}
in which $y$ is the correct image class corresponding to $\{\textbf{\textit{v}}_{i}\}_{i=1}^{K\times R}$, $\textbf{\textit{m}}_{y}$ is the prototype of class $y$, $\textbf{\textit{m}}_{r,i}$ is the prototype from the incorrect classes that is closest to $\textbf{\textit{v}}_{i}$.
\begin{align}
	L_{MFA\_DCE}=-\sum\nolimits_{i=1}^{K\times R}log\frac{e^{-||\textbf{\textit{v}}_{i}-\textbf{\textit{m}}_{y}||_{2}^{2}}}{\sum\nolimits_{j=1}^{C}e^{-||\textbf{\textit{v}}_{i}-\textbf{\textit{m}}_{j}||_{2}^{2}}}
\end{align}
At inference time, the predicted class $p$ is:
\begin{align}
	p=\mathop{\arg\min}_{j}(\sum\nolimits_{i=1}^{K\times R}||\textbf{\textit{v}}_{i}-\textbf{\textit{m}}_{j}||_{2}^{2}).
\end{align}

\renewcommand{\arraystretch}{1.5}
\begin{table}[H]  
  \centering  
  \fontsize{8.5}{9}\selectfont  
    \begin{tabular}{|p{14mm}<{\centering}||p{12mm}<{\centering}p{12mm}<{\centering}p{10mm}<{\centering}p{10mm}<{\centering}|}
    % \toprule[2pt]
    \hline  
    \multirow{2}{*}{\textbf{Model}}&  
    \multicolumn{4}{c|}{\textbf{Datasets}}\cr\cline{2-5}  
    &Imagenette&Indoor67&Action40&AID\cr  
    \hline
    % \midrule[1.2pt]
    \hline  
    Pro$_{8}$\_GAP&80.26\%&66.46\%&50.28\%&86.74\%\cr\hline  
    Pro$_{8}$\_MFA&86.48\%&73.77\%&58.05\%&94.06\%\cr\hline
    \hline  
    Pro$_{16}$\_GAP&81.83\%&68.11\%&52.25\%&87.28\%\cr\hline  
    Pro$_{16}$\_MFA&87.74\%&74.86\%&60.12\%&94.44\%\cr\hline
    % \midrule[1.2pt]
    \hline  
    Pro$_{32}$\_GAP&82.44\%&68.58\%&52.75\%&87.86\%\cr\hline  
    Pro$_{32}$\_MFA&88.75\%&75.30\%&60.50\%&94.60\%\cr\hline  
    % \midrule[1.2pt]
    \hline 
    Pro$_{64}$\_GAP&83.24\%&68.87\%&53.21\%&88.67\%\cr\hline  
    Pro$_{64}$\_MFA&89.10\%&75.63\%&60.89\%&95.55\%\cr\hline
    % \midrule[1.2pt]
    \hline  
    Pro$_{128}$\_GAP&83.08\%&69.34\%&53.52\%&88.20\%\cr\hline  
    Pro$_{128}$\_MFA&89.00\%&76.09\%&61.05\%&95.43\%\cr\hline
    % \bottomrule[2pt]
    \end{tabular} 
	\caption{Comparison between GAP and MV-FeaAug ($K=50$) in validation accuracy when combined with prototype classifier head trained with MCE loss. The used CNN backbone is VGG16. Pro$_{l}$\_GAP and Pro$_{l}$\_MFA respectively denotes GAP-based and MV-FeaAug-based model that uses the prototype classifier head where the dimension of the prototype is $l$.}
\label{tab:performance_with_prototype_cls_head_MCE}
\end{table}

\renewcommand{\arraystretch}{1.5}
\begin{table}[H]  
  \centering  
  \fontsize{8.5}{9}\selectfont  
    \begin{tabular}{|p{14mm}<{\centering}||p{12mm}<{\centering}p{12mm}<{\centering}p{10mm}<{\centering}p{10mm}<{\centering}|}
    % \toprule[2pt]
    \hline  
    \multirow{2}{*}{\textbf{Model}}&  
    \multicolumn{4}{c|}{\textbf{Datasets}}\cr\cline{2-5}  
    &Imagenette&Indoor67&Action40&AID\cr  
    \hline
    % \midrule[1.2pt]
    \hline  
    Pro$_{8}$\_GAP&80.70\%&67.63\%&51.34\%&87.20\%\cr\hline  
    Pro$_{8}$\_MFA&86.56\%&74.40\%&58.78\%&95.28\%\cr\hline
    \hline  
    Pro$_{16}$\_GAP&82.36\%&68.38\%&52.45\%&88.14\%\cr\hline  
    Pro$_{16}$\_MFA&87.95\%&74.88\%&60.11\%&95.55\%\cr\hline
    % \midrule[1.2pt]
    \hline  
    Pro$_{32}$\_GAP&82.84\%&69.00\%&53.24\%&88.89\%\cr\hline  
    Pro$_{32}$\_MFA&88.79\%&75.95\%&60.80\%&95.73\%\cr\hline  
    % \midrule[1.2pt]
    \hline 
    Pro$_{64}$\_GAP&83.75\%&69.66\%&54.08\%&89.36\%\cr\hline  
    Pro$_{64}$\_MFA&89.87\%&76.24\%&61.20\%&96.51\%\cr\hline
    % \midrule[1.2pt]
    \hline  
    Pro$_{128}$\_GAP&83.54\%&69.89\%&53.73\%&88.54\%\cr\hline  
    Pro$_{128}$\_MFA&89.45\%&76.40\%&61.05\%&95.66\%\cr\hline
    % \bottomrule[2pt]
    \end{tabular} 
	\caption{Comparison between GAP and MV-FeaAug ($K=50$) in validation accuracy when combined with prototype classifier head trained with DCE loss. The used CNN backbone is VGG16. Pro$_{l}$\_GAP and Pro$_{l}$\_MFA respectively denotes GAP-based and MV-FeaAug-based model that uses the prototype classifier head where the dimension of the prototype is $l$.}
\label{tab:performance_with_prototype_cls_head_DCE}
\end{table}

Validation accuracies achieved on different datasets using the prototype classifier head trained with MCE loss and DCE loss are reported in Table \ref{tab:performance_with_prototype_cls_head_MCE} and Table \ref{tab:performance_with_prototype_cls_head_DCE} respectively. We observe that in the case of prototype classifier head, our MV-FeaAug still remarkably improves classification performance compared with GAP. Besides, we also study the impact of the dimension of the prototype vector to model performance, our tested prototype dimension values are 8, 16, 32, 64, 128, among which it is shown in Table \ref{tab:performance_with_prototype_cls_head_MCE} and Table \ref{tab:performance_with_prototype_cls_head_DCE} that 64-dimensional prototypes lead to best performance. 

\section{Ablation study}
In this part, we investigate the contributing factors of our MV-FeaAug for improving model performance. Since our method relates to both feature augmentation and attention mechanism, we build two models to test classification performance without feature augmentation and without AdaCAM respectively, which we call \textbf{W/O FeaAug} and \textbf{W/O AdaCAM}. For both two models, we use VGG16 backbone and single-FC-layer softmax classifier head.

For model \textbf{W/O FeaAug}, we do not perform multi-view local feature sampling for feature augmentation, but use the attention map produced by AdaCAM to spatially highlight useful features and suppress irrelevant features on $\textbf{\textit{F}}=\{F_{k}\}_{k=1}^{K}$, the final output feature maps of the CNN backbone. Concretely, based on the attention map $A$ generated via our AdaCAM, i.e., $A$=AdaCAM($\textbf{\textit{F}}$), we build an attention mask $\textit{M}$:
\begin{align}
	\textit{M}(x,y)=\frac{2}{1+e^{-\eta A(x,y)}},
	\label{attention mask}
\end{align}
where the hyperparameter $\eta$ controls the steepness of the mapping function. With Equation \ref{attention mask}, the attention map $A$ is mapped into the range (0, 2) to selectively amplify or minify local features on $\textbf{\textit{F}}$, leading to the enhanced feature maps $\hat{\textbf{\textit{F}}}$:
\begin{align}
	\hat{\textbf{\textit{F}}}=\textbf{\textit{F}} \otimes M,
\end{align}
in which $M$ and $\textbf{\textit{F}}$ share the same spatial size, and $\otimes$ is multiplication with broadcast. We set $\eta=10$ in Equation \ref{attention mask} to make the mapping function steeper than the standard sigmoid function, such that more elements in $M$ are shrinked to 0 in order to further weaken useless features on  $\textbf{\textit{F}}$. The enhanced feature maps $\hat{\textbf{\textit{F}}}$ are vectorized through GAP and fed to the subsequent classifier head for classification.

For model \textbf{W/O AdaCAM}, we do not use AdaCAM to guide attention to class-related local regions, but sample 49 evenly-spaced anchor points on the 14$\times$14 feature maps $\textbf{\textit{F}}$ for feature augmentation. We still keep the same region size list $L_{R}$=[3,5,7,9], and still use the ensembled predictions of the augmented multi-view local features as the final prediction of the entire image.
\renewcommand{\arraystretch}{1.8}
\begin{table}[H]  
  \centering  
  \fontsize{8}{9.5}\selectfont  
    \begin{tabular}{|p{12mm}<{\centering}|p{9mm}<{\centering}|p{10mm}<{\centering}|p{14mm}<{\centering}|p{10mm}<{\centering}|}
    % \toprule[1.5pt]
    \hline  
    \textbf{Datasets}&\textbf{GAP}&\textbf{W/O FeaAug}&\textbf{W/O AdaCAM}&\textbf{MV-FeaAug}\cr\hline
    % \midrule[1.2pt]
    \hline  
    Imagenette&82.04\%&82.87\%&85.28\%&88.45\%\cr\hline 
    Caltech101&70.42\%&70.93\%&73.85\%&76.05\%\cr\hline
    Event8&80.00\%&80.67\%&84.06\%&85.75\%\cr\hline 
    Action40&52.19\%&53.05\%&57.24\%&60.36\%\cr\hline 
    UCMLU&90.95\%&91.25\%&93.48\%&96.67\%\cr\hline 
    Indoor67&68.25\%&68.84\%&72.55\%&75.27\%\cr\hline
    AID&87.61\%&88.35\%&92.14\%&94.55\%\cr\hline
    \end{tabular}  
    \caption{Ablation study of the contributing factors of our method for classification performance (validation accuracy). We set $K=49$ in our MV-FeaAug module. The used CNN backbone is VGG16, and the used classifier head is single-FC-layer softmax classifier head.}
    \label{tab:ablation_study}
\end{table} 

Ablation study results are reported in Table \ref{tab:ablation_study}, we observe that both $\textbf{W/O FeaAug}$ and $\textbf{W/O AdaCAM}$ outperform the GAP-based baseline, which indicates that both feature augmentation and local region attention provided by AdaCAM are useful to model performance improvements. However, the model $\textbf{W/O FeaAug}$ only boosts validation accuracy moderately while the model $\textbf{W/O AdaCAM}$, by contrast, brings much more noticeable performance gains, which demonstrates that feature augmentation is a more important factor in our MV-FeaAug module, namely, the performance gains mainly come from sampling diverse multi-view local features for ensemble learning. On the other hand, our full model MV-FeaAug consistently performs betten than the model $\textbf{W/O AdaCAM}$, which reflects that attending to class-discriminative local regions of feature maps via our AdaCAM is also contributive to model performance, since it helps to make the augmented multi-view local features more class-related.

\section{More visualization of the sampled multi-view local regions}

\begin{figure*}[t]
\centering
\includegraphics[width=6.6in]{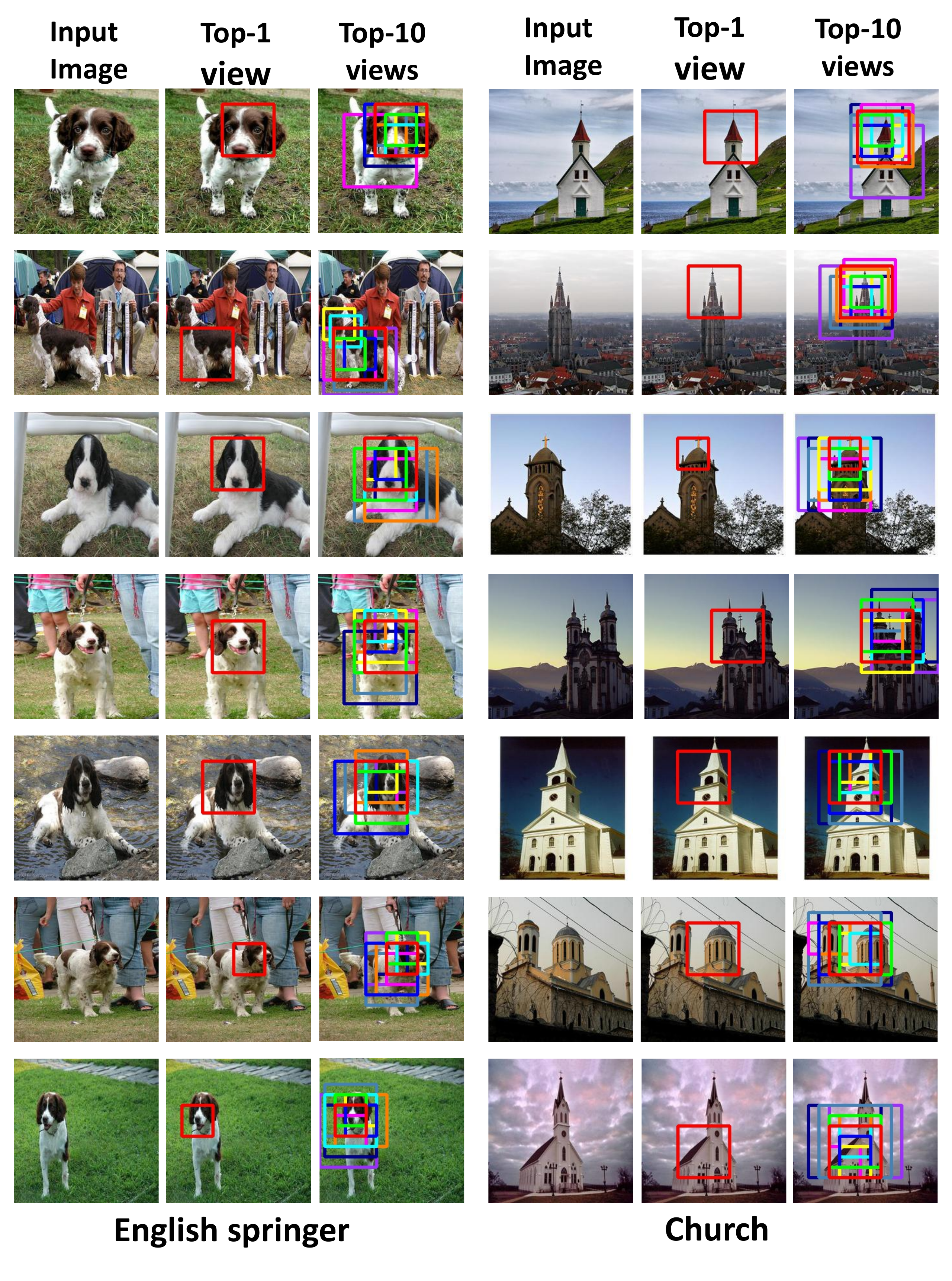}
\caption{Visualization of the top-1 and top-10 local regions that correspond to local features with the highest prediction confidence of the corresponding image class. Results are evaluated on the "English springer" and "church" categories of the Imagenette validation set.}
\label{top_k_views_1}
\end{figure*}

\begin{figure*}[t]
\centering
\includegraphics[width=6.6in]{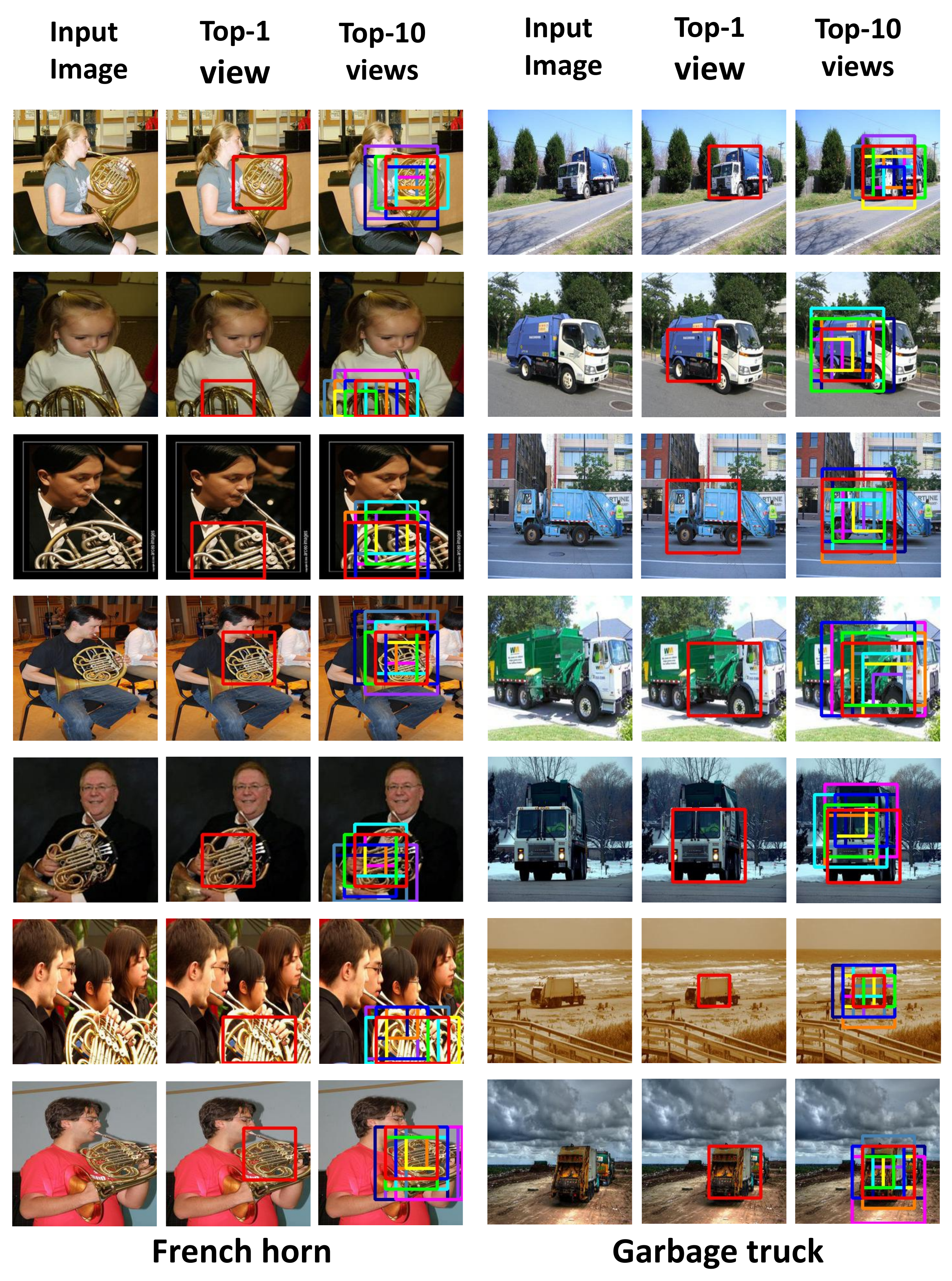}
\caption{Visualization of the top-1 and top-10 local regions that correspond to local features with the highest prediction confidence of the corresponding image class. Results are evaluated on the "French horn" and "Garbage truck" categories of the Imagenette validation set.}
\label{top_k_views_2}
\end{figure*}

Based on VGG16\_MFA, we select from all the sampled $K\times R$ multi-view local regions top-k ones that correspond to local features with the highest prediction confidence of the corresponding image class in the main classifier head, and map these top-k regions from feature scale to image scale as a visualization of our MV-FeaAug. Results of top-1 and top-10 local regions evaluated on images of the Imagenette validation set are shown in Figure \ref{top_k_views_1} and Figure \ref{top_k_views_2}, we see that the augmented multi-view local features correspond to local regions that basically gather around the objects of the corresponding class, which demonstrates effectiveness of our method in extracting rich and class-related local features as a kind of feature-level data augmentation.

\bibliographystyle{named}
\bibliography{ijcai21}

\begin{thebibliography}{}

\bibitem[\protect\citeauthoryear{Cao \bgroup \em et al.\egroup }{2019}]{gcnet}
Yue Cao, Jiarui Xu, Stephen Lin, Fangyun Wei, and Han Hu.
\newblock Gcnet: Non-local networks meet squeeze-excitation networks and
  beyond.
\newblock In {\em Proceedings of the IEEE International Conference on Computer
  Vision Workshops}, pages 0--0, 2019.

\bibitem[\protect\citeauthoryear{DeVries and Taylor}{2017}]{cutout}
Terrance DeVries and Graham~W Taylor.
\newblock Improved regularization of convolutional neural networks with cutout.
\newblock {\em arXiv preprint arXiv:1708.04552}, 2017.

\bibitem[\protect\citeauthoryear{Duygulu \bgroup \em et al.\egroup
  }{2002}]{corel5k}
Pinar Duygulu, Kobus Barnard, Joao~FG de~Freitas, and David~A Forsyth.
\newblock Object recognition as machine translation: Learning a lexicon for a
  fixed image vocabulary.
\newblock In {\em European conference on computer vision}, pages 97--112.
  Springer, 2002.

\bibitem[\protect\citeauthoryear{Fei-Fei \bgroup \em et al.\egroup
  }{2004}]{caltech101}
Li~Fei-Fei, Rob Fergus, and Pietro Perona.
\newblock Learning generative visual models from few training examples: An
  incremental bayesian approach tested on 101 object categories.
\newblock In {\em 2004 conference on computer vision and pattern recognition
  workshop}, pages 178--178. IEEE, 2004.

\bibitem[\protect\citeauthoryear{He \bgroup \em et al.\egroup }{2016}]{resnet}
Kaiming He, Xiangyu Zhang, Shaoqing Ren, and Jian Sun.
\newblock Deep residual learning for image recognition.
\newblock In {\em Proceedings of the IEEE conference on computer vision and
  pattern recognition}, pages 770--778, 2016.

\bibitem[\protect\citeauthoryear{Howard}{}]{imagenette}
Jeremy Howard.
\newblock imagenette, 2019.
\newblock {\em URL https://github.com/fastai/imagenette}.

\bibitem[\protect\citeauthoryear{Howard \bgroup \em et al.\egroup
  }{2017}]{mobilenets}
Andrew~G Howard, Menglong Zhu, Bo~Chen, Dmitry Kalenichenko, Weijun Wang,
  Tobias Weyand, Marco Andreetto, and Hartwig Adam.
\newblock Mobilenets: Efficient convolutional neural networks for mobile vision
  applications.
\newblock {\em arXiv preprint arXiv:1704.04861}, 2017.

\bibitem[\protect\citeauthoryear{Hu \bgroup \em et al.\egroup }{2018}]{SENet}
Jie Hu, Li~Shen, and Gang Sun.
\newblock Squeeze-and-excitation networks.
\newblock In {\em Proceedings of the IEEE conference on computer vision and
  pattern recognition}, pages 7132--7141, 2018.

\bibitem[\protect\citeauthoryear{Huang \bgroup \em et al.\egroup
  }{2017}]{huang2017densely}
Gao Huang, Zhuang Liu, Laurens Van Der~Maaten, and Kilian~Q Weinberger.
\newblock Densely connected convolutional networks.
\newblock In {\em Proceedings of the IEEE conference on computer vision and
  pattern recognition}, pages 4700--4708, 2017.

\bibitem[\protect\citeauthoryear{Iandola \bgroup \em et al.\egroup
  }{2016}]{squeezenet}
Forrest~N Iandola, Song Han, Matthew~W Moskewicz, Khalid Ashraf, William~J
  Dally, and Kurt Keutzer.
\newblock Squeezenet: Alexnet-level accuracy with 50x fewer parameters and< 0.5
  mb model size.
\newblock {\em arXiv preprint arXiv:1602.07360}, 2016.

\bibitem[\protect\citeauthoryear{Lazebnik \bgroup \em et al.\egroup
  }{2006}]{scene15}
Svetlana Lazebnik, Cordelia Schmid, and Jean Ponce.
\newblock Beyond bags of features: Spatial pyramid matching for recognizing
  natural scene categories.
\newblock In {\em 2006 IEEE Computer Society Conference on Computer Vision and
  Pattern Recognition (CVPR'06)}, volume~2, pages 2169--2178. IEEE, 2006.

\bibitem[\protect\citeauthoryear{Li and Fei-Fei}{2007a}]{event8}
Li-Jia Li and Li~Fei-Fei.
\newblock What, where and who? classifying events by scene and object
  recognition.
\newblock In {\em 2007 IEEE 11th international conference on computer vision},
  pages 1--8. IEEE, 2007.

\bibitem[\protect\citeauthoryear{Li and Fei-Fei}{2007b}]{UCMerced}
Li-Jia Li and Li~Fei-Fei.
\newblock What, where and who? classifying events by scene and object
  recognition.
\newblock In {\em 2007 IEEE 11th international conference on computer vision},
  pages 1--8. IEEE, 2007.

\bibitem[\protect\citeauthoryear{Li \bgroup \em et al.\egroup }{2019}]{SKNet}
Xiang Li, Wenhai Wang, Xiaolin Hu, and Jian Yang.
\newblock Selective kernel networks.
\newblock In {\em Proceedings of the IEEE conference on computer vision and
  pattern recognition}, pages 510--519, 2019.

\bibitem[\protect\citeauthoryear{Lin \bgroup \em et al.\egroup }{2013}]{NIN}
Min Lin, Qiang Chen, and Shuicheng Yan.
\newblock Network in network.
\newblock {\em arXiv preprint arXiv:1312.4400}, 2013.

\bibitem[\protect\citeauthoryear{Quattoni and Torralba}{2009}]{indoor67}
Ariadna Quattoni and Antonio Torralba.
\newblock Recognizing indoor scenes.
\newblock In {\em 2009 IEEE Conference on Computer Vision and Pattern
  Recognition}, pages 413--420. IEEE, 2009.

\bibitem[\protect\citeauthoryear{Selvaraju \bgroup \em et al.\egroup
  }{2017}]{Grad-CAM}
Ramprasaath~R Selvaraju, Michael Cogswell, Abhishek Das, Ramakrishna Vedantam,
  Devi Parikh, and Dhruv Batra.
\newblock Grad-cam: Visual explanations from deep networks via gradient-based
  localization.
\newblock In {\em Proceedings of the IEEE international conference on computer
  vision}, pages 618--626, 2017.

\bibitem[\protect\citeauthoryear{Simonyan and Zisserman}{2014}]{vggnet}
Karen Simonyan and Andrew Zisserman.
\newblock Very deep convolutional networks for large-scale image recognition.
\newblock {\em arXiv preprint arXiv:1409.1556}, 2014.

\bibitem[\protect\citeauthoryear{Szegedy \bgroup \em et al.\egroup
  }{2015}]{googlenet}
Christian Szegedy, Wei Liu, Yangqing Jia, Pierre Sermanet, Scott Reed, Dragomir
  Anguelov, Dumitru Erhan, Vincent Vanhoucke, and Andrew Rabinovich.
\newblock Going deeper with convolutions.
\newblock In {\em Proceedings of the IEEE conference on computer vision and
  pattern recognition}, pages 1--9, 2015.

\bibitem[\protect\citeauthoryear{Wang \bgroup \em et al.\egroup
  }{2017}]{wang2017residual}
Fei Wang, Mengqing Jiang, Chen Qian, Shuo Yang, Cheng Li, Honggang Zhang,
  Xiaogang Wang, and Xiaoou Tang.
\newblock Residual attention network for image classification.
\newblock In {\em Proceedings of the IEEE conference on computer vision and
  pattern recognition}, pages 3156--3164, 2017.

\bibitem[\protect\citeauthoryear{Woo \bgroup \em et al.\egroup }{2018}]{cbam}
Sanghyun Woo, Jongchan Park, Joon-Young Lee, and In~So~Kweon.
\newblock Cbam: Convolutional block attention module.
\newblock In {\em Proceedings of the European conference on computer vision
  (ECCV)}, pages 3--19, 2018.

\bibitem[\protect\citeauthoryear{Xia \bgroup \em et al.\egroup }{2017}]{AID}
Gui-Song Xia, Jingwen Hu, Fan Hu, Baoguang Shi, Xiang Bai, Yanfei Zhong,
  Liangpei Zhang, and Xiaoqiang Lu.
\newblock Aid: A benchmark data set for performance evaluation of aerial scene
  classification.
\newblock {\em IEEE Transactions on Geoscience and Remote Sensing},
  55(7):3965--3981, 2017.

\bibitem[\protect\citeauthoryear{Xie \bgroup \em et al.\egroup
  }{2017}]{ResNeXt}
Saining Xie, Ross Girshick, Piotr Doll{\'a}r, Zhuowen Tu, and Kaiming He.
\newblock Aggregated residual transformations for deep neural networks.
\newblock In {\em Proceedings of the IEEE conference on computer vision and
  pattern recognition}, pages 1492--1500, 2017.

\bibitem[\protect\citeauthoryear{Yang \bgroup \em et al.\egroup
  }{2018a}]{yang2018robust}
Hong-Ming Yang, Xu-Yao Zhang, Fei Yin, and Cheng-Lin Liu.
\newblock Robust classification with convolutional prototype learning.
\newblock In {\em Proceedings of the IEEE Conference on Computer Vision and
  Pattern Recognition}, pages 3474--3482, 2018.

\bibitem[\protect\citeauthoryear{Yang \bgroup \em et al.\egroup
  }{2018b}]{prototype}
Hong-Ming Yang, Xu-Yao Zhang, Fei Yin, and Cheng-Lin Liu.
\newblock Robust classification with convolutional prototype learning.
\newblock In {\em Proceedings of the IEEE Conference on Computer Vision and
  Pattern Recognition}, pages 3474--3482, 2018.

\bibitem[\protect\citeauthoryear{Yao \bgroup \em et al.\egroup
  }{2011}]{action40}
Bangpeng Yao, Xiaoye Jiang, Aditya Khosla, Andy~Lai Lin, Leonidas Guibas, and
  Li~Fei-Fei.
\newblock Human action recognition by learning bases of action attributes and
  parts.
\newblock In {\em 2011 International conference on computer vision}, pages
  1331--1338. IEEE, 2011.

\bibitem[\protect\citeauthoryear{Yun \bgroup \em et al.\egroup }{2019}]{cutmix}
Sangdoo Yun, Dongyoon Han, Seong~Joon Oh, Sanghyuk Chun, Junsuk Choe, and
  Youngjoon Yoo.
\newblock Cutmix: Regularization strategy to train strong classifiers with
  localizable features.
\newblock In {\em Proceedings of the IEEE International Conference on Computer
  Vision}, pages 6023--6032, 2019.

\bibitem[\protect\citeauthoryear{Zagoruyko and Komodakis}{2016}]{wideresnet}
Sergey Zagoruyko and Nikos Komodakis.
\newblock Wide residual networks.
\newblock {\em arXiv preprint arXiv:1605.07146}, 2016.

\bibitem[\protect\citeauthoryear{Zhang \bgroup \em et al.\egroup
  }{2017}]{mixup}
Hongyi Zhang, Moustapha Cisse, Yann~N Dauphin, and David Lopez-Paz.
\newblock mixup: Beyond empirical risk minimization.
\newblock {\em arXiv preprint arXiv:1710.09412}, 2017.

\bibitem[\protect\citeauthoryear{Zhang \bgroup \em et al.\egroup
  }{2020}]{zhang2020resnest}
Hang Zhang, Chongruo Wu, Zhongyue Zhang, Yi~Zhu, Zhi Zhang, Haibin Lin, Yue
  Sun, Tong He, Jonas Mueller, R~Manmatha, et~al.
\newblock Resnest: Split-attention networks.
\newblock {\em arXiv preprint arXiv:2004.08955}, 2020.

\bibitem[\protect\citeauthoryear{Zhou \bgroup \em et al.\egroup }{2016}]{CAM}
Bolei Zhou, Aditya Khosla, Agata Lapedriza, Aude Oliva, and Antonio Torralba.
\newblock Learning deep features for discriminative localization.
\newblock In {\em Proceedings of the IEEE conference on computer vision and
  pattern recognition}, pages 2921--2929, 2016.

\bibitem[\protect\citeauthoryear{Zou \bgroup \em et al.\egroup }{2015}]{RSSCN7}
Qin Zou, Lihao Ni, Tong Zhang, and Qian Wang.
\newblock Deep learning based feature selection for remote sensing scene
  classification.
\newblock {\em IEEE Geoscience and Remote Sensing Letters}, 12(11):2321--2325,
  2015.

\end{thebibliography}

\end{document}